\documentclass[runningheads]{llncs}

\usepackage{eccv}

\usepackage{eccvabbrv}

\usepackage{graphicx}
\usepackage{booktabs}

\usepackage[accsupp]{axessibility}  

\usepackage{bbm}

\usepackage{hyperref}

\usepackage{orcidlink}

\usepackage{colortbl}
\newcommand{\CC}[0]{\cellcolor{LightGray}}
\definecolor{LightGray}{gray}{0.93}

\begin{document}

\title{Improving Adversarial Robustness via Activation Amplification and Attenuation}

\titlerunning{A3: Activation Amplification and Attenuation}

\author{
	Taïga GONÇALVES\inst{1}\orcidlink{0009-0004-1019-2031} \and
	Yongsong HUANG\inst{1}\orcidlink{0000-0003-3114-9206} \and
	Tomo MIYAZAKI\inst{1}\orcidlink{0000-0001-5205-0542} \and
	Shinichiro OMACHI\inst{1}\orcidlink{0000-0001-7706-9995}
}
\authorrunning{Gonçalves~et al.}

\institute{Graduate School of Engineering, Tohoku University, Japan 
\email{goncalves.taiga.teo.t6@dc.tohoku.ac.jp},
\email{huang.yongsong.c5@tohoku.ac.jp},
\email{tomo@tohoku.ac.jp},
\email{shinichiro.omachi.b5@tohoku.ac.jp} 	
}

\maketitle

\begin{abstract}
	The existence of adversarial attacks is often attributed to the presence of non-robust features in neural networks. While prior defenses reduce their impact via pruning, masking, or feature recalibration, we instead propose to jointly learn to amplify and attenuate these signals through a simple activation scaling mechanism. To this end, we introduce \textbf{Activation Amplification and Attenuation (A3)}, a lightweight plug-in module that enhances adversarial robustness with minimal modifications of the activations. A3 dynamically rescales the activations using a learnable mask and a scaling factor derived from the original activation magnitudes. The influence of adversarial perturbations can be amplified or attenuated using the same learnable parameters by simply flipping the sign of the scaling operation. The amplified signals serve as negative references to construct novel contrastive and ranking loss functions. Experimental analysis shows that learning to degrade the predictions in amplification mode simultaneously improves adversarial robustness in attenuation mode. Moreover, A3 relies on only a small number of learnable parameters, with most of its behavior being determined by the scaling mechanism rather than additional network capacity. Extensive experiments demonstrate that integrating A3 into different backbones, datasets, and training methods consistently improves adversarial robustness while introducing negligible computational and memory overhead compared to existing plug-in modules. Code is available at: \href{https://github.com/tgoncalv/A3}{https://github.com/tgoncalv/A3}.
\end{abstract}

\begin{figure}[tb]
	\hfill
	\centering
	\begin{subfigure}{0.44\linewidth}
		\includegraphics[width=\linewidth]{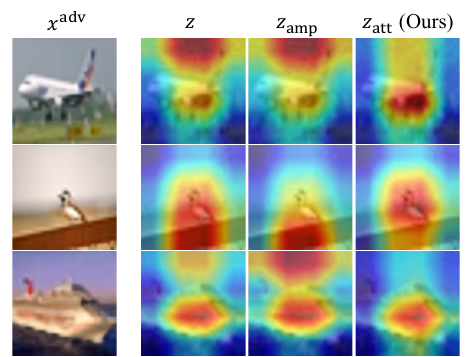}
		\caption{CIFAR-10}
		\label{fig:grad_cif10}
	\end{subfigure}
	\hfill
	\rule[.35cm]{0.1mm}{3.86cm} 
	\hfill
	\centering
	\begin{subfigure}{0.44\linewidth}
		\includegraphics[width=\linewidth]{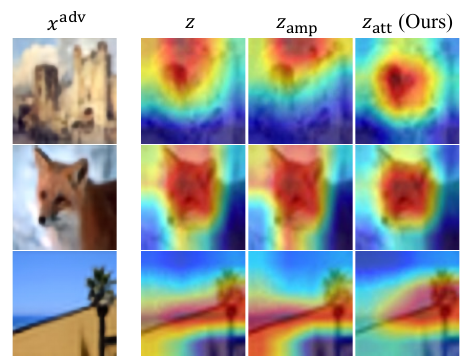}
		\caption{CIFAR-100}
		\label{fig:grad_cif100}
	\end{subfigure}
	\hfill
	\caption{
		\textbf{Grad-CAM attention maps~\cite{GradCAM} on adversarial examples.} We compute Grad-CAM at \textit{block}~4 of vanilla ResNet-18 (before inserting A3) using the activations $z$, and compare it with Grad-CAM obtained from A3 in amplification ($z_\text{amp}$) and attenuation ($z_\text{att}$) modes. The amplification mode tends to emphasize class-irrelevant regions (\eg background sky rather than the airplane/ship), while the attenuation mode reduces the contribution of these regions and focuses more on class-relevant areas. During training, we use $z_\text{amp}$ as a negative reference to construct novel contrastive and ranking losses, which contributes to the robustness improvements observed in the attenuation mode.
	}
	\label{fig:gradcam}
\end{figure}

\section{Introduction}
\label{sec:intro}

The increasing success of Deep Neural Networks (DNNs) across a wide range of applications raises concerns about their security and reliability. In particular, adversarial examples~\cite{AdvAtt_Pandas, AdvAtt_PGD} add imperceptible perturbations specifically designed to mislead the models' predictions. To address this vulnerability, numerous works have proposed defense methods to improve the adversarial robustness of DNNs, together with standardized evaluation protocols for assessing their robustness under various types of attacks~\cite{Adv_RobustBench, Adv_AutoAttack}.

Current defense techniques can be broadly divided into two categories. The first group focuses on training strategies~\cite{Adv_TRADES, adv_MART, wasedaRethinkingInvarianceRegularization2025}, which use adversarial examples to introduce new objective functions and regularization terms without modifying the network architecture. The second group directly modifies internal feature representations~\cite{Adv_FSR,Adv_EMFF,Adv_FPCM} by inserting lightweight plug-in modules into the backbone, making the network inherently more robust.

In practice, these plug-in modules generally follow the same design principles. \textbf{(i) Adaptability:} the module should be compatible with different model architectures and training strategies. \textbf{(ii) Efficiency:} the computational and memory overhead should be minimal. \textbf{(iii) Effectiveness:} the accuracy on adversarial examples must improve compared to the vanilla backbone. \textbf{(iv) Reliability:} the module must remain differentiable and interpretable to avoid reliability issues such as gradient obfuscation~\cite{Adv_Adaptive, Adv_Obfuscation}.

Some plug-in modules compute a mask to identify and suppress non-robust features~\cite{Adv_CAS,Adv_CIFS}. More recent methods~\cite{Adv_FSR,Adv_FTA2C} argued that completely discarding non-robust features is suboptimal, as they may still contain useful information. Therefore, they propose to separate the features into robust and non-robust components, and recalibrate the latter using additional multi-layer perceptrons (MLPs). While effective, these recalibration blocks mainly introduce extra learnable parameters whose contributions to robustness are less interpretable, since the architecture of these MLPs is fundamentally not much different from the other components of the original backbone.

To this end, we propose \textbf{Activation Amplification and Attenuation (A3)}, a lightweight plug-in module designed to strengthen internal feature representations via a simple rescaling mechanism. Our design is inspired by recent advances in Out-of-Distribution (OOD) detection~\cite{OOD_ASH, OOD_SCALE}, which found that appropriately scaling the original activations can significantly improve the separability between In-Distribution (ID) and OOD samples. Since adversarial examples can be viewed as a form of OOD data, we propose a scaling operation specifically designed to improve adversarial robustness. A3 computes a sample-wise scaling factor using a learnable masking function and the distribution of the original activation magnitudes, so that different channels are rescaled with different strengths. This controlled rescaling encourages a clearer separation between useful and less useful predictive signals, thereby focusing robustness improvements on useful cues, as shown in~\cref{fig:gradcam}. Moreover, unlike traditional approaches that only suppress or down-weight the influence of non-robust features, A3 supports both amplification and attenuation of these signals using the same learnable parameters by simply flipping the sign of the scaling operation. We then exploit the amplified signals as training-time negative references to define novel contrastive and ranking loss functions. Our contributions can be summarized as follows:

\begin{itemize}
	\item We propose \textbf{A3}, a lightweight plug-in module that can be integrated into different architectures to improve adversarial robustness with minimal computational and memory overhead.
	\item We design a simple scaling mechanism that can either amplify or attenuate activation patterns to focus robustness improvements on useful predictive signals. We further introduce novel contrastive and ranking loss functions that exploit the amplified activations as negative references to enhance the attenuation mode.
	\item We show that A3 consistently improves adversarial robustness across models, datasets, and adversarial training methods compared to prior plug-in defense modules. Extensive experiments and ablation studies highlight the importance of both amplification and attenuation modes.
\end{itemize}

\section{Related Work}

\subsection{Adversarial Training}

The most widely used strategy to enhance model robustness is adversarial training (AT)~\cite{AdvAtt_Pandas,AdvAtt_PGD}, formulated as the following min-max optimization problem:

\begin{equation}\label{eq:minmax}
	\min_{\theta} \mathbb{E}_{(x,y) \sim D} \left[ \max_{\delta \in S} L(f_\theta(x+\delta), y) \right],
\end{equation}
where $f_\theta$ is a neural network with parameters $\theta$, $L$ is a loss function, $D$ is the dataset, and $S$ is the set of allowed perturbations, typically defined as an $\ell_p$-norm constraint around the clean input $x$. The inner maximization aims to find the optimal perturbation $\delta$ that maximizes the training loss under the constraint set $S$ to induce misclassification. The perturbation is generally obtained using gradient-based attacks such as the Fast Gradient Sign Method (FGSM)~\cite{AdvAtt_Pandas} or the Projected Gradient Descent (PGD)~\cite{AdvAtt_PGD}. The outer minimization then updates the model parameters $\theta$ to minimize the loss on these adversarial examples.

Many extensions have been proposed to improve the standard AT strategy. TRADES~\cite{Adv_TRADES} and MART~\cite{adv_MART} combine both clean and adversarial examples in the outer minimization process to improve model robustness while maintaining high accuracy on clean data. AGR~\cite{Adv_AGR} further refines this trade-off by comparing clean and adversarial examples in the gradient space rather than the input space. Other works strengthen the inner maximization to generate more effective adversarial examples that better guide the model to learn robust features. For example, LAS-AT~\cite{Adv_LAS-AT} dynamically adjusts the attack strength on a per-sample basis, while CFA~\cite{Adv_CFA} adapts attack configurations in a class-wise manner. Finally, Yu \etal~\cite{Adv_SGLS} show that using hard ground-truth labels often lead to robust overfitting and propose Self-Guided Label Smoothing (SGLS), which uses self-distillation to generate soft labels that improve inference-time robustness.

\subsection{Robust Feature Manipulation}

Another line of work introduces plug-in modules that can be integrated into different architectures to improve the robustness of intermediate features. Xie \etal~\cite{xieFeatureDenoisingImproving2019} propose a simple feature denoising block based on non-local filtering and a $1 \times 1$ convolution to reduce adversarial noise while preserving meaningful signals. FPCM~\cite{Adv_FPCM} reconfigures the low and high-frequency feature components based on the observation that adversarial perturbations often manifest in the high-frequency domain. EMFF~\cite{Adv_EMFF} leverages evidential deep learning to estimate feature uncertainty across multiple layers and uses this information to refine the learned representations.

Recent works also focus on identifying non-robust features and suppressing their influence. CAS~\cite{Adv_CAS} and CIFS~\cite{Adv_CIFS} observed that such features are often concentrated in specific channels and proposed modules that identify and suppress these channels on a per-sample basis. Later, Kim \etal~\cite{Adv_FSR} argued that completely discarding non-robust features is a suboptimal solution as they may still contain useful information. Therefore, they introduced the Feature Separation and Recalibration (FSR) module, which separates robust and non-robust components and recalibrates the latter to reduce adversarial noise while preserving useful cues. FTA2C~\cite{Adv_FTA2C} further refines this recalibration process by enhancing feature alignment between clean and adversarial examples.

\subsection{Activation Scaling in OOD Detection}

Out-of-distribution (OOD) detection aims to identify inputs that do not belong to the training distribution. Many approaches quantify how likely an input is to be OOD using scoring functions $S(z):\mathcal{Z} \rightarrow \mathbb{R}$~\cite{OOD_ODIN, OOD_Mahalanobis, OOD_Energy} based on the original features or logits. A widely used function is the energy score~\cite{OOD_Energy}:

\begin{equation}\label{eq:energy_OOD}
	S_\mathit{energy}(z) = -T \cdot \log \sum_{i=1}^{C} e^{z_i/T},
\end{equation}
where $T$ is a temperature hyperparameter, $C$ is the number of channels, and $z$ is a $C$-dimensional vector at the penultimate layer of the model. This function assigns lower energy scores for in-distribution (ID) samples compared to OOD samples, which can then be separated using a simple threshold value.

Recent works have shown that rescaling part of the activations before applying the energy-based scoring can significantly improve the energy gap between ID and OOD data. ASH~\cite{OOD_ASH} proposed to use a top-k operation to selectively prune low-magnitude activations while amplifying the remaining ones. Then, SCALE~\cite{OOD_SCALE} demonstrated that the pruning step yields suboptimal results and that scaling all activations with a carefully designed scaling factor leads to better performance. Later, LTS~\cite{OOD_LTS} and AdaSCALE~\cite{OOD_AdaSCALE} further enhanced the performance of SCALE while preserving the core scaling principle.

\begin{figure}[tb]
	\centering
	\includegraphics[width=0.7\linewidth]{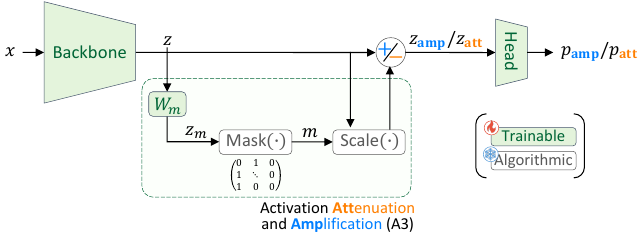}
	\caption{
		\textbf{Overview of the proposed A3 module.} A3 is a plug-in module that can be integrated into various models. The activations $z$ are processed using a small learnable projection $W_m$ and a Gumbel-Softmax $\text{Mask}(\cdot)$ operator to produce a channel-wise mask. The resulting mask and activation magnitudes are then passed to $\text{Scale}(\cdot)$ to compute a rescaling factor used to either amplify ($+$) or attenuate ($-$) the activations depending on the sign of the operation. During training, $z_\text{amp}$ serves as a negative reference while $z_\text{att}$ serves as a positive reference to define novel contrastive and ranking loss functions. During inference, the final predictions are obtained solely from $z_\text{att}$.
	}
	\label{fig:a3m}
\end{figure}

\section{Method}

\subsection{Activation Scaling}

\subsubsection{Preliminaries.}

OOD detection methods based on activation scaling~\cite{OOD_ASH,OOD_LTS,OOD_SCALE,OOD_AdaSCALE} employ a top-k operation to compute a scaling factor defined as:

\begin{equation}\label{OOD_Scale}
	\text{Scale}_\text{OOD}(z, p) = exp\left(\frac{\sum_{i} z_i}{\sum_{z_i > P_p(z)} z_i}\right),
\end{equation}
where $P_p(z)$ denotes the p-th percentile of the activations $z$. These activations are typically extracted from the penultimate layer of the model, after a ReLU operation. Therefore, all the values $z_i$ are positive and thus $\text{Scale}_\text{OOD}(z, p)~\geq~1$. Then, the energy score function in \cref{eq:energy_OOD} is evaluated by replacing the original activations $z$ with the scaled activations $z \cdot \text{Scale}_\text{OOD}(z, p)$. This simple rescaling process results in a larger energy gap between ID and OOD data, thereby facilitating their separation using a threshold.

Although adversarial examples can be viewed as a form of OOD data, several challenges must be addressed to adapt the scaling method for adversarial training. First, the top-k operation used in OOD methods is non-differentiable. In the context of adversarial training, this can increase the risk of gradient obfuscation~\cite{Adv_Obfuscation,Adv_Adaptive}. This vulnerability was notably highlighted for k-WTA~\cite{kWTA}, which relied on a top-k operation to suppress non-robust features but was later found to be affected by gradient obfuscation~\cite{Adv_Adaptive}. OOD detection methods are unaffected by this issue because the energy scoring functions are post-hoc methods that are only used during inference, thus do not need to guarantee differentiability. Second, the exponential formulation in \cref{OOD_Scale} produces excessively large factors, which can cause gradient saturation and numerical instability. These limitations motivate us to retain the intuition behind activation scaling while introducing a differentiable and stable method specifically designed for adversarial robustness.

\subsubsection{Proposed Approach.}

We present an overview of A3 in \cref{fig:a3m}. To rescale the activations $z$ in a controlled and differentiable manner, we first predict a channel-wise binary mask $m$ using the Gumbel-Softmax trick~\cite{GumbelSoftmax}:

\begin{equation}\label{eq:GumbelSoftmax}
	\text{Mask}(z_{m}, \tau_{m}) =
	\frac{
		e^{((\log(\sigma(z_{m})) + g_1)/\tau_{m})}}{
		e^{((\log(\sigma(z_{m})) + g_1)/\tau_{m})} + e^{((\log(1 - \sigma(z_{m})) + g_2)/\tau_{m})}
	},
\end{equation}
where $\sigma(\cdot)$ is the sigmoid function, $\tau_{m}$ is a temperature hyperparameter, and $g_1, g_2$ are \iid samples drawn from the Gumbel distribution~\cite{gumbelStatisticalTheoryExtreme1954}. The vector $z_m$ is obtained by applying Global Average Pooling (GAP) to $z$ across its spatial dimensions and a linear projection using a learnable weight matrix $W_m \in \mathbb{R}^{C \times C}$:

\begin{equation}\label{eq:zw}
	z_m = W_m \cdot \text{GAP}(z), \qquad
	\text{GAP}(z)_c = \frac{1}{H W} \sum_{h=1}^{H} \sum_{w=1}^{W} z_{c,h,w}.
\end{equation}

The Gumbel-Softmax and linear projection are employed to obtain a learnable masking operation that captures useful predictive patterns in the activations. The generated mask $m$ is then used to compute two statistical values $s_1$ and $s_2$ based on the distribution of the original activation magnitudes:

\begin{equation}\label{eq:s1s2}
	a_c =\sum_{h=1}^{H} \sum_{w=1}^{W} |z_{c,h,w}|, \qquad
	s_1 = \sum_{c=1}^{C} a_c, \qquad
	s_2 = \sum_{c=1}^{C} a_c \cdot m_c.
\end{equation}

Here, $s_1$ represents the overall magnitude of the activations, while $s_2$ captures the magnitude within the masked activations. These formulations are inspired by the scaling function in \cref{OOD_Scale}, but we replace the non-differentiable top-k selection with a learnable, differentiable mask to prevent the risk of gradient obfuscation. The learnable projection $W_m$ also allows the masking operation to adapt the scaling function to different models and datasets. Using these two statistics, we compute two scaling factors $f_m$ and $f_{1m}$ as follows:

\begin{equation}\label{eq:fm}
	f_m = \frac{\ln(1 + s_2)}{\ln(1 + s_1)}, \qquad
	f_{1m} = 1 - f_m.
\end{equation}

There are several important aspects to note about these scaling factors. First, $f_m$ and $f_{1m}$ are differentiable with respect to the activations $z$ and the mask $m$, enabling seamless end-to-end optimization. Second, the logarithmic formulation provides more stable gradient behavior compared to the exponential-based scaling function in \cref{OOD_Scale}. In particular, the derivative of the logarithm decays smoothly as input magnitudes increase, which helps mitigate gradient explosion and reduces sensitivity to large activation outliers. This stability is particularly crucial in adversarial training, where adversarial perturbations often induce large activation spikes. We also empirically verify in our ablation study that this formulation is the most effective for our task. We then combine the scaling factors with the mask to compute the following scaling function:

\begin{equation}\label{eq:ScaleNew}
	\text{Scale}(z,m) = f_m\cdot m + f_{1m}\cdot (1 - m).
\end{equation}

Finally, we modify the original activations $z$ to obtain either amplified or attenuated signals using the following operation:

\begin{equation}\label{eq:zatt_zamp}
	z_{\text{att}} = z \cdot \bigl[1 - \text{Scale}(z,m) \bigr], \qquad
	z_{\text{amp}} = z \cdot \bigl[1 + \text{Scale}(z,m) \bigr].
\end{equation}

Unlike the OOD scaling function in \cref{OOD_Scale}, our operation is designed such that the scaling factor $\text{Scale}(z,m)$ is always bounded within $[0, 1]$. In fact, since $s_2 \leq s_1$ by definition, it follows that $f_m \in [0, 1]$ and $f_{1m} \in [0, 1]$. This property ensures that the scaling factors remain bounded, preventing extreme amplification or attenuation of the activations, while also preserving a distribution close to the original activations. During inference, the final predictions are computed solely from the attenuated activations $z_\text{att}$. The amplified activations $z_\text{amp}$ are only used during training to serve as negative references for the contrastive and ranking loss functions described in the following section.

\subsection{Objective Functions}

\subsubsection{Cross-Entropy Ranking Loss.}

To encourage the model to attenuate the influence of adversarial attacks on useful predictive signals, we introduce a ranking loss that enforces the attenuated activations to yield a lower cross-entropy (CE) loss than the amplified activations, using the following formulation:

\begin{equation}{\label{eq:loss_rank}}
	L_\text{rank} = \max(\text{CE}(\hat{p}_\text{att}^\text{adv}, y) - \text{CE}(\hat{p}_\text{amp}, y), 0),
\end{equation}
where $\hat{p}$ denotes the softmax probability output and $y$ is the ground-truth label. This objective serves two primary roles. First, it establishes a competitive optimization process within the module itself, since it simultaneously encourages lower error for attenuated signals and higher error for amplified signals. Second, the hinge loss formulation helps prevent the model from overfitting to specific adversarial examples. We validate the contributions of both the competitive optimization and the hinge formulation in our ablation study.

\subsubsection{Contrastive Logit Loss.}

Since the attenuation mode aims to reduce the influence of harmful activations, the resulting logits should ideally be invariant to adversarial perturbations. To encourage this behavior, we introduce a contrastive logit loss defined as follows:

\begin{equation}{\label{eq:loss_triplet}}
	L_\text{cl} = -\log\frac{e^{(sim(l_\text{att}^\text{adv}, l_\text{att})/\tau_\text{cl})}}{e^{(sim(l_\text{att}^\text{adv}, l_\text{att})/\tau_\text{cl})} + e^{(sim(l_\text{att}^\text{adv}, l_\text{amp})/\tau_\text{cl})}},
\end{equation}
where $l \in \mathbb{R}^k$ denotes the logit vector for $k$ classes produced by the final classifier head, $sim(\cdot, \cdot)$ is the cosine similarity function, and $\tau_\text{cl}$ is a temperature hyperparameter controlling the sharpness of the similarity distribution. This objective simultaneously minimizes the distance between clean and adversarial logits obtained from the attenuation mode, while also enhancing the separability between the attenuated and amplified logits.

\textbf{Overall Loss.} Our final objective function integrates the standard adversarial training loss with our proposed component losses:
\begin{equation}{\label{eq:loss_total}}
	L = L_\text{main} + \lambda_\text{rank} L_\text{rank} + \lambda_\text{cl} L_\text{cl},
\end{equation}
where $L_\text{main}$ is the original training loss of the selected adversarial training strategy (\eg, AT~\cite{AdvAtt_PGD}, TRADES~\cite{Adv_TRADES}, MART~\cite{adv_MART}), and $\lambda_\text{rank}$, $\lambda_\text{cl}$ are hyperparameters that control the contribution of our proposed ranking and contrastive losses, respectively.

\begin{table}[tb]
	\centering
	\caption{
		Robust accuracy (\%) of \textbf{ResNet-18} on CIFAR-10/100 datasets under various attacks. The "clean" column denotes accuracy without attacks. We evaluate the vanilla model, our A3, and other plug-in defenses under different training strategies (AT, TRADES, MART). For each strategy, the best results are in \textbf{bold} and the second-best are \underline{underlined}.
	}
	\setlength{\tabcolsep}{2pt}
	\resizebox{1.00\linewidth}{!}{
		\begin{tabular}{p{4.3cm}|c|cccccc||c|cccccc}
			\hline
			\multicolumn{1}{c|}{\textbf{Method}}          & \multicolumn{7}{c||}{\textbf{CIFAR-10}} & \multicolumn{7}{c}{\textbf{CIFAR-100}}                                                                                                                                                                                                                                                                \\
			\hline
			\centering \textbf{ResNet-18}                 & \textbf{Clean}                          & \textbf{FGSM}                          & \textbf{PGD-20}    & \textbf{PGD-100}   & \textbf{C\&W}      & \textbf{Ens.}      & \textbf{AA}        & \textbf{Clean}     & \textbf{FGSM}      & \textbf{PGD-20}    & \textbf{PGD-100}      & \textbf{C\&W}      & \textbf{Ens.}      & \textbf{AA}        \\
			\hline
			\hline
			\textbf{AT (Baseline)}                        & 85.04                                   & 56.96                                  & 49.12              & 47.51              & 48.19              & 46.05              & 44.28              & 59.26              & 29.83              & 25.17              & 24.16                 & 24.28              & 22.85              & 21.26              \\
			+ CAS \textit{(ICLR 2021)} \cite{Adv_CAS}     & 82.45                                   & 56.81                                  & 50.02              & 48.93              & \underline{50.20}  & 46.97              & 43.40              & 58.48              & 28.92              & 24.83              & 23.84                 & 24.82              & 22.55              & 20.12              \\
			+ CIFS \textit{(ICML 2021)} \cite{Adv_CIFS}   & 80.62                                   & 57.34                                  & 51.30              & 49.45              & 49.06              & 47.55              & 43.65              & 56.64              & 29.11              & 25.14              & 24.21                 & 24.35              & 22.98              & 20.41              \\
			+ FSR \textit{(CVPR 2023)} \cite{Adv_FSR}     & 82.78                                   & 57.51                                  & 51.45              & 50.05              & 49.24              & \underline{47.67}  & 45.92              & 58.84              & 29.10              & 25.24              & 24.36                 & 24.51              & 23.05              & 21.94              \\
			+ FPCM \textit{(ICCV 2023)} \cite{Adv_FPCM}   & \underline{85.51}                       & \underline{58.07}                      & 49.92              & 48.22              & 49.44              & 47.25              & 46.14              & \underline{60.05}  & \underline{32.24}  & \underline{28.49}  & \underline{27.45}     & \underline{27.05}  & \underline{25.38}  & \underline{24.25}  \\
			+ RiFT \textit{(ICCV 2023)} \cite{Adv_RiFT}   & \textbf{85.57}                          & 57.92                                  & 50.79              & 49.19              & 49.45              & 47.59              & \underline{46.21}  & \textbf{60.17}     & 30.43              & 26.08              & 25.34                 & 25.26              & 23.89              & 22.75              \\
			+ EMFF \textit{(T-PAMI 2025)} \cite{Adv_EMFF} & 81.82                                   & 57.08                                  & 51.04              & 49.63              & 48.58              & 47.25              & 45.84              & 57.17              & 31.49              & 27.79              & 27.02                 & 26.40              & 25.22              & 24.02              \\
			+ FTA2C \textit{(NN 2026)} \cite{Adv_FTA2C}   & 82.35                                   & 57.50                                  & \underline{52.03}  & \underline{50.88}  & 48.70              & 47.14              & 45.42              & 56.12              & 30.75              & 26.69              & 26.02                 & 24.85              & 23.32              & 21.82              \\
			\CC \textbf{+ A3 (Ours)}                      & \CC 84.37                               & \CC \textbf{62.43}                     & \CC \textbf{58.66} & \CC \textbf{57.01} & \CC \textbf{52.33} & \CC \textbf{51.04} & \CC \textbf{47.28} & \CC 57.50          & \CC \textbf{33.89} & \CC \textbf{32.34} & \CC \textbf{31.78}    & \CC \textbf{27.44} & \CC \textbf{26.52} & \CC \textbf{24.36} \\
			\hline
			\hline
			\textbf{TRADES (Baseline)}                    & \underline{84.02}                       & 57.83                                  & 51.59              & 50.48              & 48.93              & 48.26              & 46.98              & 59.02              & 31.29              & 27.44              & 26.98                 & 24.42              & 24.07              & 23.05              \\
			+ CAS \textit{(ICLR 2021)} \cite{Adv_CAS}     & 83.19                                   & 57.02                                  & 51.24              & 50.31              & 49.05              & 47.99              & 45.12              & 57.41              & 30.22              & 27.41              & 26.59                 & 24.90              & 24.26              & 22.08              \\
			+ CIFS \textit{(ICML 2021)} \cite{Adv_CIFS}   & 81.34                                   & 57.40                                  & 51.58              & 50.47              & 48.95              & 48.23              & 45.39              & 57.68              & 30.30              & 27.52              & 26.94                 & 24.48              & 23.97              & 22.19              \\
			+ FSR \textit{(CVPR 2023)} \cite{Adv_FSR}     & 83.46                                   & 57.49                                  & 51.67              & 50.78              & 49.09              & 48.39              & 47.50              & 57.63              & 30.39              & 27.57              & 27.01                 & 24.53              & 24.13              & 23.35              \\
			+ FPCM \textit{(ICCV 2023)} \cite{Adv_FPCM}   & 83.73                                   & 58.41                                  & 53.14              & \underline{52.27}  & \underline{50.61}  & \underline{49.68}  & 48.67              & 58.98              & \underline{31.95}  & \underline{29.40}  & \textbf{29.05}        & \underline{26.03}  & \underline{25.45}  & \underline{24.50}  \\
			+ RiFT \textit{(ICCV 2023)} \cite{Adv_RiFT}   & \textbf{84.75}                          & \underline{58.48}                      & 52.47              & 51.26              & 50.01              & 49.35              & 48.24              & \textbf{60.73}     & 31.74              & 28.51              & 27.92                 & 25.42              & 24.95              & 23.91              \\
			+ EMFF \textit{(T-PAMI 2025)} \cite{Adv_EMFF} & 83.40                                   & 57.96                                  & 52.09              & 50.67              & 48.91              & 48.23              & 47.13              & \underline{59.35}  & \textbf{32.43}     & 29.26              & 28.79                 & 25.68              & 25.28              & 24.49              \\
			+ FTA2C \textit{(NN 2026)} \cite{Adv_FTA2C}   & 83.97                                   & 58.19                                  & \underline{52.99}  & 51.90              & 50.27              & 49.60              & \underline{48.68}  & 57.88              & 31.47              & 28.31              & 27.91                 & 25.66              & 24.91              & 24.00              \\
			\CC \textbf{+ A3 (Ours)}                      & \CC 83.77                               & \CC \textbf{58.50}                     & \CC \textbf{55.10} & \CC \textbf{54.26} & \CC \textbf{51.52} & \CC \textbf{50.97} & \CC \textbf{49.27} & \CC 57.36          & \CC 31.71          & \CC \textbf{29.61} & \CC \underline{28.81} & \CC \textbf{26.35} & \CC \textbf{25.80} & \CC \textbf{24.81} \\
			\hline
			\hline
			\textbf{MART (Baseline)}                      & 81.92                                   & 57.98                                  & 52.12              & 51.08              & 48.25              & 47.56              & 45.88              & \underline{57.92}  & 31.36              & 27.41              & 26.67                 & 24.96              & 24.38              & 22.19              \\
			+ CAS \textit{(ICLR 2021)} \cite{Adv_CAS}     & 80.89                                   & 57.73                                  & 51.17              & 49.88              & \underline{49.13}  & 47.16              & 43.75              & 55.24              & 30.27              & 26.68              & 25.89                 & 25.11              & 23.31              & 21.12              \\
			+ CIFS \textit{(ICML 2021)} \cite{Adv_CIFS}   & 81.01                                   & 58.19                                  & 52.24              & 50.67              & 48.62              & 47.66              & 44.06              & 54.14              & 30.40              & 26.82              & 26.17                 & 24.84              & 23.53              & 21.44              \\
			+ FSR \textit{(CVPR 2023)} \cite{Adv_FSR}     & \underline{83.21}                       & 58.38                                  & 52.50              & 51.02              & 48.95              & 47.78              & 46.16              & 56.45              & 30.55              & 27.01              & 26.28                 & 25.27              & 24.15              & 23.02              \\
			+ FPCM \textit{(ICCV 2023)} \cite{Adv_FPCM}   & 80.92                                   & 55.86                                  & 49.37              & 47.92              & 45.65              & 44.86              & 43.46              & 55.08              & 28.14              & 25.50              & 24.94                 & 22.54              & 22.19              & 21.24              \\
			+ RiFT \textit{(ICCV 2023)} \cite{Adv_RiFT}   & \textbf{83.53}                          & \underline{58.73}                      & \underline{52.87}  & \underline{51.43}  & 48.73              & \underline{47.93}  & \underline{46.31}  & 56.15              & \underline{32.32}  & \underline{29.25}  & \underline{28.78}     & \underline{25.96}  & \underline{25.26}  & 23.92              \\
			+ EMFF \textit{(T-PAMI 2025)} \cite{Adv_EMFF} & 82.13                                   & 57.87                                  & 51.59              & 50.16              & 48.14              & 46.99              & 45.61              & 55.87              & 31.47              & 28.45              & 27.88                 & 25.81              & 25.21              & \underline{24.14}  \\
			+ FTA2C \textit{(NN 2026)} \cite{Adv_FTA2C}   & 82.16                                   & 58.00                                  & 52.64              & 51.37              & 48.32              & 47.34              & 45.69              & 55.38              & 30.85              & 27.45              & 26.65                 & 24.21              & 23.27              & 21.80              \\
			\CC \textbf{+ A3 (Ours)}                      & \CC 82.69                               & \CC \textbf{62.83}                     & \CC \textbf{59.05} & \CC \textbf{57.05} & \CC \textbf{51.70} & \CC \textbf{50.86} & \CC \textbf{47.29} & \CC \textbf{58.86} & \CC \textbf{33.01} & \CC \textbf{29.79} & \CC \textbf{28.70}    & \CC \textbf{27.28} & \CC \textbf{26.01} & \CC \textbf{24.24} \\
			\hline
		\end{tabular}
	}
	\label{tab:results_resnet18}
\end{table}
\begin{table}[tb]
	\centering
	\caption{
		Robust accuracy (\%) of \textbf{WideResNet-34-10} on CIFAR-10/100 datasets under various attacks. The "clean" column denotes accuracy without attacks. We evaluate the vanilla model, our A3, and other plug-in defenses under different training strategies (AT, TRADES, MART). For each strategy, the best results are in \textbf{bold} and the second-best are \underline{underlined}.
	}
	\setlength{\tabcolsep}{2pt}
	\resizebox{1.00\linewidth}{!}{
		\begin{tabular}{p{4.3cm}|c|cccccc||c|cccccc}
			\hline

			\multicolumn{1}{c|}{\textbf{WideResNet-34-10}} & \multicolumn{7}{c||}{\textbf{CIFAR-10}} & \multicolumn{7}{c}{\textbf{CIFAR-100}}                                                                                                                                                                                                                                                               \\
			\hline
			\centering \textbf{Method}                     & \textbf{Clean}                          & \textbf{FGSM}                          & \textbf{PGD-20}    & \textbf{PGD-100}   & \textbf{C\&W}      & \textbf{Ens.}      & \textbf{AA}        & \textbf{Clean}    & \textbf{FGSM}         & \textbf{PGD-20}    & \textbf{PGD-100}   & \textbf{C\&W}      & \textbf{Ens.}      & \textbf{AA}        \\
			\hline
			\hline
			\textbf{AT (Baseline)}                         & \underline{87.57}                       & 60.12                                  & 51.58              & 49.83              & 51.60              & 49.27              & 48.18              & \underline{62.71} & 31.59                 & 26.53              & 25.48              & 26.55              & 24.69              & 23.70              \\
			+ CAS \textit{(ICLR 2021)} \cite{Adv_CAS}      & 86.43                                   & 60.16                                  & 51.67              & 50.02              & 51.43              & 49.23              & 46.62              & 61.98             & 31.78                 & 26.68              & 25.98              & 26.84              & 24.52              & 22.37              \\
			+ CIFS \textit{(ICML 2021)} \cite{Adv_CIFS}    & 85.96                                   & 60.91                                  & 51.78              & 50.18              & 52.07              & 49.58              & 46.59              & 61.75             & 31.92                 & 26.84              & 25.74              & 26.52              & 24.70              & 22.63              \\
			+ FSR \textit{(CVPR 2023)} \cite{Adv_FSR}      & 87.07                                   & 61.13                                  & 53.81              & \underline{51.97}  & \underline{52.41}  & \underline{50.65}  & 48.80              & 61.88             & 30.92                 & 25.11              & 23.61              & 25.64              & 23.33              & 22.28              \\
			+ FPCM \textit{(ICCV 2023)} \cite{Adv_FPCM}    & 87.13                                   & 58.70                                  & 50.48              & 48.54              & 50.12              & 48.31              & 47.36              & 61.41             & 30.41                 & 26.07              & 25.14              & 26.02              & 24.56              & 23.99              \\
			+ RiFT \textit{(ICCV 2023)} \cite{Adv_RiFT}    & \textbf{87.85}                          & 60.83                                  & 52.32              & 50.38              & 52.05              & 49.84              & 48.74              & \textbf{63.00}    & 31.71                 & 26.65              & 25.50              & 26.55              & 24.94              & 24.23              \\
			+ EMFF \textit{(T-PAMI 2025)} \cite{Adv_EMFF}  & 86.10                                   & \underline{62.29}                      & \underline{53.92}  & 51.75              & 51.78              & 50.38              & \underline{49.10}  & 60.33             & \underline{34.22}     & \underline{30.05}  & \underline{29.07}  & \underline{28.59}  & \underline{27.33}  & \underline{26.22}  \\
			+ FTA2C \textit{(NN 2026)} \cite{Adv_FTA2C}    & 86.36                                   & 60.44                                  & 52.70              & 51.07              & 52.22              & 49.73              & 48.10              & 61.07             & 30.71                 & 25.84              & 24.53              & 25.42              & 23.21              & 22.07              \\
			\CC \textbf{+ A3 (Ours)}                       & \CC 86.64                               & \CC \textbf{63.01}                     & \CC \textbf{58.33} & \CC \textbf{56.72} & \CC \textbf{55.32} & \CC \textbf{53.82} & \CC \textbf{51.62} & \CC 62.50         & \CC \textbf{34.64}    & \CC \textbf{31.22} & \CC \textbf{30.23} & \CC \textbf{30.51} & \CC \textbf{28.98} & \CC \textbf{27.68} \\
			\hline
			\hline
			\textbf{TRADES (Baseline)}                     & 86.39                                   & 61.14                                  & 52.69              & 50.86              & 52.12              & 50.25              & 49.33              & 61.88             & 32.50                 & 28.31              & 27.62              & 27.04              & 26.32              & 25.51              \\
			+ CAS \textit{(ICLR 2021)} \cite{Adv_CAS}      & 86.30                                   & 61.62                                  & 52.84              & 51.12              & 52.81              & 50.56              & 47.61              & 60.74             & 32.03                 & 28.09              & 27.53              & 27.31              & 26.29              & 24.49              \\
			+ CIFS \textit{(ICML 2021)} \cite{Adv_CIFS}    & 86.08                                   & 61.27                                  & 52.98              & 51.27              & 52.34              & 50.71              & 47.96              & 60.18             & 32.27                 & 28.34              & 27.70              & 27.14              & 26.11              & 24.60              \\
			+ FSR \textit{(CVPR 2023)} \cite{Adv_FSR}      & \underline{86.92}                       & \underline{62.86}                      & 54.24              & 51.97              & \underline{52.98}  & \underline{51.10}  & 50.08              & 59.13             & 32.70                 & 29.40              & 28.87              & 27.43              & 26.89              & 26.22              \\
			+ FPCM \textit{(ICCV 2023)} \cite{Adv_FPCM}    & 85.83                                   & 58.96                                  & 51.76              & 50.12              & 50.85              & 49.36              & 48.54              & 58.31             & 31.57                 & 28.23              & 27.53              & 26.45              & 25.80              & 24.97              \\
			+ RiFT \textit{(ICCV 2023)} \cite{Adv_RiFT}    & \textbf{87.29}                          & 61.92                                  & 53.25              & 51.26              & 52.54              & 50.58              & 49.64              & \underline{62.93} & 33.44                 & 29.36              & 28.35              & 27.84              & 27.08              & 26.21              \\
			+ EMFF \textit{(T-PAMI 2025)} \cite{Adv_EMFF}  & 86.51                                   & 62.41                                  & \underline{54.30}  & \underline{52.24}  & 52.01              & 50.47              & 49.66              & \textbf{63.66}    & \textbf{34.47}        & \underline{29.78}  & \underline{28.96}  & \underline{27.97}  & \underline{27.22}  & \underline{26.24}  \\
			+ FTA2C \textit{(NN 2026)} \cite{Adv_FTA2C}    & 84.44                                   & 62.08                                  & 54.38              & 52.15              & 52.36              & 50.95              & \underline{50.17}  & 57.23             & 32.57                 & 29.61              & 28.71              & 27.32              & 26.43              & 26.20              \\
			\CC \textbf{+ A3 (Ours)}                       & \CC 85.46                               & \CC \textbf{65.34}                     & \CC \textbf{58.19} & \CC \textbf{55.12} & \CC \textbf{54.84} & \CC \textbf{53.28} & \CC \textbf{51.41} & \CC 59.78         & \CC \underline{33.87} & \CC \textbf{31.45} & \CC \textbf{30.61} & \CC \textbf{28.26} & \CC \textbf{27.66} & \CC \textbf{26.46} \\
			\hline
			\hline
			\textbf{MART (Baseline)}                       & \underline{86.79}                       & 60.78                                  & 53.05              & 51.08              & 51.00              & 49.62              & 48.51              & \underline{61.39} & 31.35                 & 26.97              & 25.85              & 25.82              & 24.74              & 23.96              \\
			+ CAS \textit{(ICLR 2021)} \cite{Adv_CAS}      & 85.75                                   & 61.32                                  & 53.21              & 51.22              & 51.57              & 50.20              & 47.27              & 60.38             & 30.51                 & 26.53              & 24.98              & 25.70              & 24.32              & 22.68              \\
			+ CIFS \textit{(ICML 2021)} \cite{Adv_CIFS}    & 86.36                                   & 61.67                                  & 53.44              & 51.31              & 51.35              & 50.39              & 47.16              & 59.79             & 30.57                 & 26.85              & 25.20              & 25.66              & 24.49              & 22.90              \\
			+ FSR \textit{(CVPR 2023)} \cite{Adv_FSR}      & 86.31                                   & 61.60                                  & 54.29              & 52.50              & 51.75              & 50.53              & 49.15              & 56.64             & 29.49                 & 24.60              & 23.53              & 24.45              & 22.89              & 22.16              \\
			+ FPCM \textit{(ICCV 2023)} \cite{Adv_FPCM}    & 86.41                                   & 61.69                                  & 54.14              & 52.17              & 52.05              & 50.87              & 49.36              & 60.42             & 31.77                 & 27.53              & 26.69              & 26.71              & 25.60              & 24.70              \\
			+ RiFT \textit{(ICCV 2023)} \cite{Adv_RiFT}    & \textbf{87.29}                          & 61.92                                  & 53.25              & 51.26              & \underline{52.54}  & 50.58              & 49.64              & \textbf{63.00}    & 31.71                 & 26.65              & 25.50              & 26.55              & 24.94              & 24.23              \\
			+ EMFF \textit{(T-PAMI 2025)} \cite{Adv_EMFF}  & 85.92                                   & \underline{63.11}                      & \underline{55.60}  & \underline{54.45}  & 52.51              & \underline{51.42}  & \underline{49.94}  & 61.04             & \underline{33.53}     & \underline{29.27}  & \underline{28.33}  & \underline{27.98}  & \underline{26.93}  & \underline{25.88}  \\
			+ FTA2C \textit{(NN 2026)} \cite{Adv_FTA2C}    & 84.93                                   & 61.30                                  & 54.09              & 52.29              & 51.18              & 50.57              & 49.18              & 57.32             & 29.62                 & 24.83              & 23.99              & 24.98              & 23.49              & 22.53              \\
			\CC \textbf{+ A3 (Ours)}                       & \CC 86.65                               & \CC \textbf{63.61}                     & \CC \textbf{58.36} & \CC \textbf{56.00} & \CC \textbf{54.54} & \CC \textbf{53.32} & \CC \textbf{50.59} & \CC 60.55         & \CC \textbf{36.19}    & \CC \textbf{33.94} & \CC \textbf{33.28} & \CC \textbf{30.90} & \CC \textbf{29.87} & \CC \textbf{28.22} \\
			\hline
		\end{tabular}
	}
	\label{tab:results_wideresnet3410}
\end{table}

\section{Experiments}

\subsection{Experimental Setup}


\textbf{Training Details.} We conduct our experiments on CIFAR-10/100~\cite{dataset_CIFAR} and Tiny ImageNet~\cite{dataset_ImageNet} datasets using ResNet-18~\cite{cls_ResNet} and WideResNet-34-10~\cite{WideResNet} as backbone architectures. We consider three representative adversarial training methods: AT~\cite{AdvAtt_PGD}, TRADES~\cite{Adv_TRADES} and MART~\cite{adv_MART}. For all methods, we employ PGD-10 attack~\cite{AdvAtt_PGD} with a maximum perturbation $\epsilon=8/255$ and step size $\epsilon/4$. We train all models for 100 epochs with a batch size of 128 using SGD optimizer (momentum 0.9, weight decay $5 \times 10^{-4}$). The initial learning rate is set to 0.1 with a decay factor of 10 at epochs 75 and 90. These settings are consistent with prior works~\cite{Adv_FSR,Adv_FTA2C,Adv_EMFF} to ensure a fair comparison. We insert the A3 module after \textit{block}~4 in ResNet-18 and after \textit{block}~3 in WideResNet-34-10. Further justification for these locations is provided in the supplementary material. Finally, we set the hyperparameters to $\tau_{m} = 0.1$, $\tau_\text{cl} = 10$, $\lambda_\text{rank} = 1$ and $\lambda_\text{cl} = 5$. The choice of these hyperparameters is discussed in the ablation study.

\begin{table}[tb]
	\centering
	\caption{
		Robust accuracy (\%) of \textbf{ResNet-18} and \textbf{WideResNet-34-10} on Tiny Imagenet dataset under various attacks. The "clean" column denotes accuracy without attacks. We evaluate the vanilla model, our A3, and other plug-in defenses under different training strategies (AT, TRADES, MART). For each strategy, the best results are in \textbf{bold} and the second-best are \underline{underlined}.
	}
	\setlength{\tabcolsep}{2pt}
	\resizebox{1.00\linewidth}{!}{
		\begin{tabular}{p{4.3cm}|c|cccccc||c|cccccc}
			\hline

			\multicolumn{1}{c|}{\textbf{Tiny ImageNet}}   & \multicolumn{7}{c||}{\textbf{ResNet-18}} & \multicolumn{7}{c}{\textbf{WideResNet-34-10}}                                                                                                                                                                                                                                                                             \\
			\hline
			\centering \textbf{Method}                    & \textbf{Clean}                           & \textbf{FGSM}                                 & \textbf{PGD-20}    & \textbf{PGD-100}      & \textbf{C\&W}      & \textbf{Ens.}      & \textbf{AA}        & \textbf{Clean}        & \textbf{FGSM}       & \textbf{PGD-20}     & \textbf{PGD-100}      & \textbf{C\&W}       & \textbf{Ens.}       & \textbf{AA}           \\
			\hline
			\hline
			\textbf{AT (Baseline)}                        & 50.33                                    & 23.83                                         & 20.34              & 19.91                 & 18.96              & 17.49              & 16.42              & 52.78                 & 26.25               & 22.84               & 22.13                 & 21.62               & 19.93               & 18.79                 \\
			+ CAS \textit{(ICLR 2021)} \cite{Adv_CAS}     & 48.85                                    & 24.16                                         & 20.56              & 20.11                 & 19.08              & 17.30              & 15.19              & 51.47                 & 26.58               & 22.97               & 22.14                 & 21.87               & 19.82               & 17.36                 \\
			+ CIFS \textit{(ICML 2021)} \cite{Adv_CIFS}   & 48.63                                    & 24.52                                         & 20.89              & 20.38                 & 18.46              & 17.49              & 15.73              & 51.72                 & 26.65               & 23.21               & 22.59                 & 21.57               & 20.01               & 17.09                 \\
			+ FSR \textit{(CVPR 2023)} \cite{Adv_FSR}     & 48.16                                    & 24.35                                         & 20.71              & 20.28                 & 19.37              & 17.52              & 16.60              & 49.64                 & 26.56               & 23.40               & 22.62                 & 21.71               & 20.10               & 19.28                 \\
			+ FPCM \textit{(ICCV 2023)} \cite{Adv_FPCM}   & 45.71                                    & \underline{24.80}                             & \underline{22.54}  & \textbf{22.30}        & \underline{19.68}  & 18.53              & 17.21              & \underline{54.92}     & \underline{27.70}   & \underline{24.30}   & \underline{23.42}     & \underline{22.45}   & \underline{20.98}   & \underline{19.80}     \\
			+ RiFT \textit{(ICCV 2023)} \cite{Adv_RiFT}   & \textbf{51.67}                           & 24.44                                         & 21.14              & 20.48                 & 19.28              & 18.04              & 16.64              & \textbf{55.03}        & 27.20               & 23.29               & 22.34                 & 22.13               & 20.42               & 19.06                 \\
			+ EMFF \textit{(T-PAMI 2025)} \cite{Adv_EMFF} & 48.49                                    & 24.16                                         & 21.69              & 21.06                 & 19.45              & \underline{18.54}  & \underline{17.43}  & 50.12                 & 26.36               & 23.88               & 23.31                 & 21.85               & 20.65               & 19.45                 \\
			+ FTA2C \textit{(NN 2026)} \cite{Adv_FTA2C}   & 48.62                                    & 22.59                                         & 20.69              & 20.50                 & 16.62              & 16.28              & 15.30              & 49.88                 & 26.28               & 23.14               & 22.49                 & 21.84               & 19.97               & 18.93                 \\
			\CC \textbf{+ A3 (Ours)}                      & \CC \underline{51.10}                    & \CC \textbf{25.46}                            & \CC \textbf{22.97} & \CC \underline{22.02} & \CC \textbf{21.32} & \CC \textbf{19.78} & \CC \textbf{18.25} & \CC 51.92             & \CC  \textbf{27.86} & \CC  \textbf{25.27} & \CC  \textbf{24.48}   & \CC  \textbf{23.32} & \CC  \textbf{21.87} & \CC  \textbf{20.43}   \\
			\hline
			\hline
			\textbf{TRADES (Baseline)}                    & 49.95                                    & 23.38                                         & 20.74              & 20.46                 & 16.83              & 16.14              & 14.80              & \underline{53.96}     & 27.08               & 24.09               & 23.88                 & 19.76               & 19.38               & 18.20                 \\
			+ CAS \textit{(ICLR 2021)} \cite{Adv_CAS}     & 48.64                                    & 23.27                                         & 20.93              & 20.77                 & 16.91              & 16.09              & 13.82              & 52.42                 & 27.64               & 24.50               & 24.00                 & 20.03               & 19.43               & 17.41                 \\
			+ CIFS \textit{(ICML 2021)} \cite{Adv_CIFS}   & 47.98                                    & 23.55                                         & 21.11              & 20.92                 & 16.76              & 16.22              & 13.97              & 52.19                 & \underline{27.79}   & 24.47               & 24.12                 & 19.84               & 19.59               & 17.28                 \\
			+ FSR \textit{(CVPR 2023)} \cite{Adv_FSR}     & 45.77                                    & 21.41                                         & 19.59              & 19.37                 & 15.69              & 15.40              & 14.47              & 50.59                 & 27.31               & 24.84               & 24.23                 & 20.29               & 19.82               & 18.39                 \\
			+ FPCM \textit{(ICCV 2023)} \cite{Adv_FPCM}   & 49.93                                    & \underline{24.29}                             & \underline{22.35}  & \underline{22.00}     & \underline{18.29}  & \underline{17.64}  & 16.08              & 53.13                 & 27.17               & 25.08               & 24.83                 & \textbf{21.25}      & 20.86               & \underline{19.75}     \\
			+ RiFT \textit{(ICCV 2023)} \cite{Adv_RiFT}   & \textbf{51.91}                           & 23.92                                         & 21.13              & 20.28                 & 16.82              & 16.37              & 13.94              & \textbf{55.79}        & 27.53               & 24.27               & 23.36                 & 20.20               & 19.68               & 18.03                 \\
			+ EMFF \textit{(T-PAMI 2025)} \cite{Adv_EMFF} & \underline{50.60}                        & 24.09                                         & 22.10              & 21.74                 & 17.37              & 17.21              & \underline{16.38}  & 53.45                 & 27.49               & \underline{25.22}   & \textbf{24.91}        & 21.11               & \textbf{20.87}      & \textbf{19.99}        \\
			+ FTA2C \textit{(NN 2026)} \cite{Adv_FTA2C}   & 48.62                                    & 22.64                                         & 20.63              & 20.46                 & 16.92              & 16.29              & 15.32              & 50.50                 & 27.65               & 24.63               & 23.93                 & 20.51               & 19.87               & 18.80                 \\
			\CC \textbf{+ A3 (Ours)}                      & \CC 48.44                                & \CC \textbf{24.95}                            & \CC \textbf{23.27} & \CC \textbf{22.86}    & \CC \textbf{19.33} & \CC \textbf{18.90} & \CC \textbf{17.73} & \CC 52.88             & \CC \textbf{27.92}  & \CC \textbf{25.28}  & \CC \underline{24.89} & \CC 20.96           & \CC 20.60           & \CC 19.07             \\
			\hline
			\hline
			\textbf{MART (Baseline)}                      & 48.05                                    & 24.48                                         & 22.21              & 21.79                 & 19.34              & 18.23              & 16.92              & 52.06                 & 27.93               & 24.86               & 24.43                 & 21.73               & 20.82               & 19.31                 \\
			+ CAS \textit{(ICLR 2021)} \cite{Adv_CAS}     & 46.64                                    & 23.75                                         & 21.79              & 20.30                 & 18.86              & 17.69              & 15.73              & 51.88                 & 28.15               & 24.87               & 24.26                 & 22.06               & 20.93               & 18.63                 \\
			+ CIFS \textit{(ICML 2021)} \cite{Adv_CIFS}   & 46.46                                    & 23.63                                         & 21.84              & 20.38                 & 18.65              & 17.56              & 15.44              & 51.49                 & 28.24               & 24.99               & 24.51                 & 21.79               & 20.72               & 18.80                 \\
			+ FSR \textit{(CVPR 2023)} \cite{Adv_FSR}     & 45.85                                    & 23.11                                         & 20.86              & 20.57                 & 17.98              & 17.33              & 16.23              & 50.50                 & 28.03               & 24.89               & 24.51                 & 21.77               & 20.91               & 19.51                 \\
			+ FPCM \textit{(ICCV 2023)} \cite{Adv_FPCM}   & 48.45                                    & 22.31                                         & 19.47              & 18.96                 & 17.62              & 16.84              & 15.88              & 53.12                 & \underline{28.57}   & \underline{26.06}   & \textbf{25.49}        & 22.76               & 21.90               & 20.37                 \\
			+ RiFT \textit{(ICCV 2023)} \cite{Adv_RiFT}   & \underline{50.20}                        & \underline{25.39}                             & 22.76              & 22.29                 & 19.74              & 18.81              & 17.40              & \textbf{53.61}        & 28.37               & 24.90               & 24.30                 & 22.20               & 21.18               & 19.76                 \\
			+ EMFF \textit{(T-PAMI 2025)} \cite{Adv_EMFF} & 46.44                                    & 24.62                                         & \underline{22.93}  & \underline{22.59}     & 19.83              & \underline{19.28}  & \underline{18.17}  & 51.88                 & 28.25               & 25.65               & 25.09                 & \underline{23.24}   & \underline{22.23}   & \textbf{21.35}        \\
			+ FTA2C \textit{(NN 2026)} \cite{Adv_FTA2C}   & 42.56                                    & 22.30                                         & 20.56              & 20.63                 & \underline{20.43}  & 17.15              & 15.53              & 50.44                 & 28.25               & 25.16               & 24.89                 & 22.30               & 21.57               & 20.11                 \\
			\CC \textbf{+ A3 (Ours)}                      & \CC \textbf{51.03}                       & \CC \textbf{26.36}                            & \CC \textbf{23.89} & \CC \textbf{23.34}    & \CC \textbf{21.78} & \CC \textbf{20.67} & \CC \textbf{18.96} & \CC \underline{53.48} & \CC \textbf{29.49}  & \CC \textbf{26.49}  & \CC \underline{25.28} & \CC \textbf{23.85}  & \CC \textbf{22.25}  & \CC \underline{20.69} \\
			\hline
		\end{tabular}
	}
	\label{tab:results_tinyimagenet}
\end{table}

\textbf{Baselines.} We compare our proposed A3 module against recent plug-in methods that are also compatible with different training methods: CAS~\cite{Adv_CAS}, CIFS~\cite{Adv_CIFS}, FPCM~\cite{Adv_FPCM}, RiFT~\cite{Adv_RiFT}, FSR~\cite{Adv_FSR}, FTA2C~\cite{Adv_FTA2C} and EMFF~\cite{Adv_EMFF}. All baselines are evaluated using the official implementations and recommended hyperparameter configurations provided by the authors.

\textbf{Evaluation Details.} Our evaluation includes clean accuracy (\ie, accuracy without attack) and robust accuracy under the following attacks: FGSM~\cite{AdvAtt_Pandas}, PGD-20, PGD-100~\cite{AdvAtt_PGD} and C\&W (with 30 steps under $\ell_\infty$ norm)~\cite{AdvAtt_CW}. Unless specified otherwise, all attacks are conducted with a maximum perturbation $\epsilon=8/255$ and a step size of $\epsilon/10$. We also report the ensemble (Ens.) accuracy, defined as the sample-wise worst-case accuracy among the following predictions: Clean, FGSM, PGD-20, PGD-100 and C\&W. Finally, we provide results under AutoAttack (AA)~\cite{Adv_AutoAttack}, which is an ensemble-based attack widely considered as the most reliable evaluation benchmark for adversarial robustness.

\subsection{Results}

We present the main results on CIFAR-10/100 in \cref{tab:results_resnet18,tab:results_wideresnet3410} and on Tiny ImageNet in \cref{tab:results_tinyimagenet}. Additional analyses such as confidence statistics and Expected Calibration Error (ECE)~\cite{ECE} are provided in the supplementary material. Unless stated otherwise, we use A3 in attenuation mode at inference time, while the amplification mode is only used during training. Overall, A3 consistently improves adversarial robustness across various models, datasets, and training methods. For example, A3 improves the ensemble accuracy (Ens.) of ResNet-18 trained with AT on CIFAR-10 by 4.99\% over the vanilla backbone, achieving the strongest performance among the compared plug-in defenses.

Although we observe a modest decrease in clean accuracy compared to the vanilla backbone, this is a common trade-off in adversarial training \cite{Adv_TRADES,Adv_RiFT}. We attribute this behavior to the relative absence of harmful activations in clean images, which makes the attenuation operation less beneficial for these samples. Nevertheless, A3 maintains a clean accuracy competitive with existing methods while ensuring better robustness against adversarial attacks, which is the primary objective of our approach. Across all three training strategies, A3 achieves the highest AA and Ens. scores without severely compromising clean accuracy.

\subsection{Ablation Studies}

\begin{table}[!tb]
	\centering
	\caption{
		\textbf{Robust accuracy (\%) of A3 under different modes.} We use ResNet-18 trained with AT on CIFAR-10 as the backbone. The attenuation mode corresponds to our default design used in all other experiments. We compare it with the amplification mode as well as two suppression modes that completely zero out either $m$ or $1-m$. The best results are highlighted in \textbf{bold}.
	}
	\setlength{\tabcolsep}{4pt}
	\resizebox{0.95\linewidth}{!}{
		\begin{tabular}{
			c
			>{\centering\arraybackslash}p{3.3cm}
			|c|cccccc}
			\hline
			\centering \textbf{Mode}                         & \textbf{A3 Output}                               & \textbf{Clean} & \textbf{FGSM}      & \textbf{PGD-20}    & \textbf{PGD-100}   & \textbf{C\&W}      & \textbf{Ens.}      & \textbf{AA}        \\
			\hline
			Suppression ($m$)                                & $z \cdot m$                                      & 32.40          & 20.85              & 19.65              & 19.26              & 16.34              & 16.11              & 15.13              \\
			Suppression ($1-m$)                              & $z \cdot (1-m)$                                  & 85.04          & 57.19              & 51.02              & 48.27              & 51.22              & 48.06              & 46.55              \\
			Amplification/$z_\text{amp}$                   & $z \cdot \bigl[1 + \text{Scale}(z,m) \bigr]$     & 84.16          & 56.20              & 50.38              & 47.81              & 50.69              & 47.36              & 45.92              \\
			Deactivated                                      & $z$                                              & \textbf{85.11} & 57.78              & 51.56              & 48.87              & 51.62              & 48.73              & 46.72              \\
			\CC Attenuation/$z_\text{att}$ \textbf{(Ours)} & \CC $z \cdot \bigl[1 - \text{Scale}(z,m) \bigr]$ & \CC 84.37      & \CC \textbf{62.43} & \CC \textbf{58.66} & \CC \textbf{57.01} & \CC \textbf{52.33} & \CC \textbf{51.04} & \CC \textbf{47.28} \\
			\hline
		\end{tabular}
	}
	\label{tab:abl_att_vs_amp}
\end{table}
\begin{table}[!t]
	\centering
	\caption{
		\textbf{Robust accuracy (\%) of A3 under different loss configurations.} We use ResNet-18 trained with AT on CIFAR-10 as the backbone. Best results are in \textbf{bold}.
	}
	\setlength{\tabcolsep}{5pt}
	\resizebox{0.90\linewidth}{!}{
		\begin{tabular}{c|c|cccccc}
			\hline
			\centering \textbf{Method}                               & \textbf{Clean}     & \textbf{FGSM}      & \textbf{PGD-20}    & \textbf{PGD-100}   & \textbf{C\&W}      & \textbf{Ens.}      & \textbf{AA}        \\
			\hline
			\CC \textbf{ResNet18 + A3 (Ours)}                        & \CC \textbf{84.37} & \CC \textbf{62.43} & \CC \textbf{58.66} & \CC \textbf{57.01} & \CC \textbf{52.33} & \CC \textbf{51.04} & \CC \textbf{47.28} \\
			w/o $L_\text{cl}$                                      & 83.57              & 61.75              & 58.02              & 56.38              & 51.35              & 50.24              & 46.70              \\
			w/o $L_\text{rank}$                                    & 83.49              & 57.03              & 51.41              & 49.03              & 49.61              & 48.11              & 45.41              \\
			$L_\text{rank}$ replaced by \cref{eq:Lrank_to_CE_dual} & 84.29              & 57.85              & 52.05              & 49.81              & 51.54              & 48.74              & 46.84              \\
			$L_\text{rank}$ replaced by \cref{eq:Lrank_to_CE_att}  & 84.05              & 58.99              & 53.50              & 51.34              & 51.39              & 49.13              & 46.73              \\
			$L_\text{rank}$ replaced by \cref{eq:Lrank_to_CE_amp}  & 11.19              & 9.67               & 9.23               & 9.17               & 9.35               & 8.95               & 7.80               \\
			\hline
		\end{tabular}
	}
	\label{tab:abl_loss}
\end{table}

\textbf{Amplification, attenuation, or suppression?} The A3 module is designed to either amplify or attenuate the influence of adversarial examples by flipping the sign of the scaling operation in \cref{eq:zatt_zamp}. Although we only need the attenuation mode during inference, we compare its performance against the amplification mode in \cref{tab:abl_att_vs_amp}. As expected, the attenuation mode achieves significantly higher accuracy on adversarial examples compared to the amplification mode. Moreover, deactivating the A3 module (\ie, replacing it with an identity function) yields higher results than the amplification mode but remains inferior to the attenuation mode. This is consistent with the intended design of A3, which aims to enhance or degrade the prediction accuracy depending on the mode. Moreover, the clean accuracy is highest when deactivating the A3 module, which is consistent with our previous analysis indicating that clean images provide fewer opportunities for the attenuation operation to improve robustness.

We also evaluated a hard suppression strategy by completely masking the activations using either $m$ or $1-m$. Our results in \cref{tab:abl_att_vs_amp} show that the suppression operations lead to a degradation in performance compared to our attenuation approach. This aligns with the findings in FSR~\cite{Adv_FSR}, which suggest that simply discarding non-robust features is a suboptimal solution.

\begin{figure}[tb]
	\centering
	\includegraphics[width=0.8\linewidth]{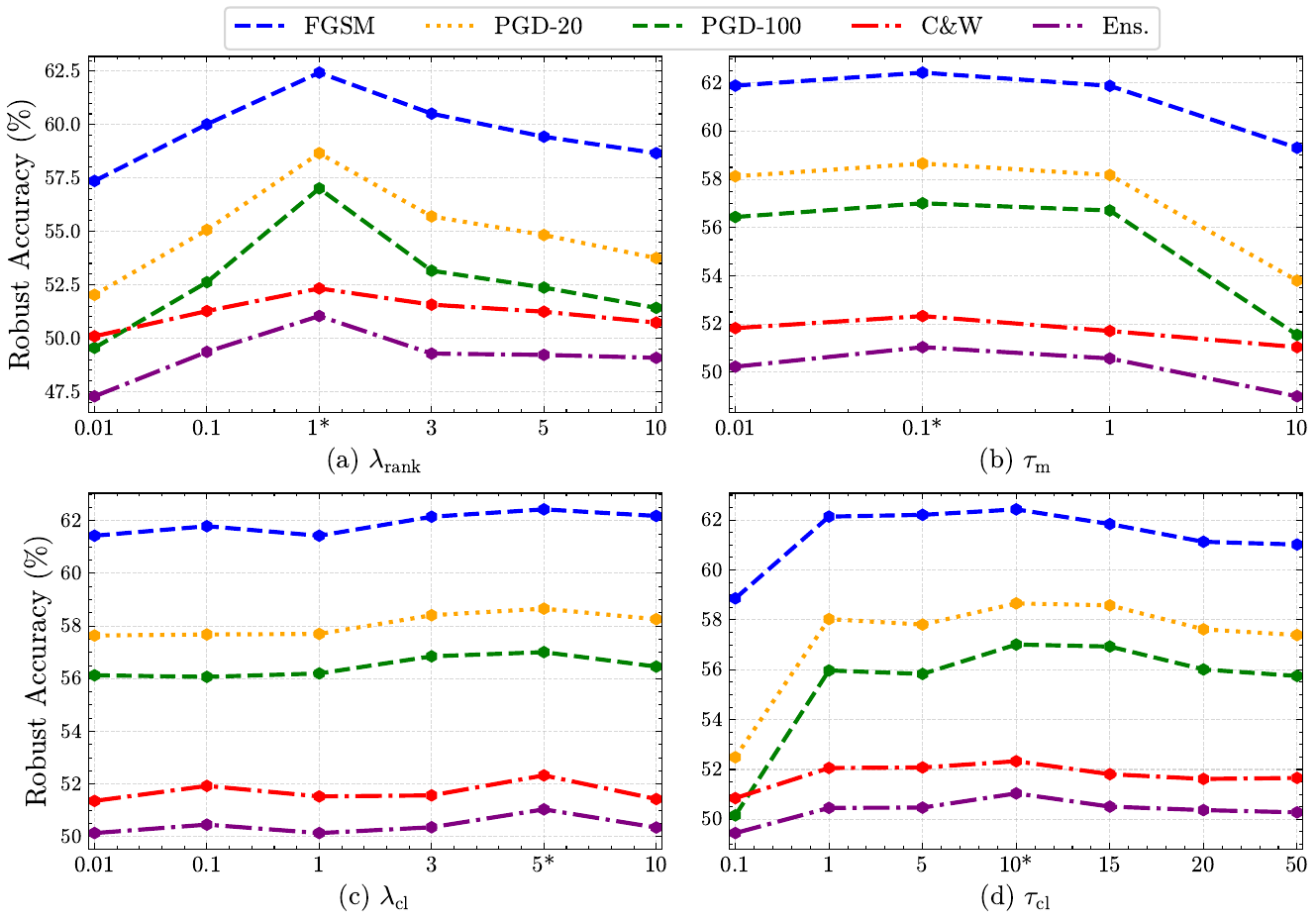}
	\caption{
		\textbf{Ablation study on hyperparameters.} We analyze the effect of different hyperparameters in our A3 module on the adversarial robustness against various attacks. We use ResNet-18 trained with AT on CIFAR-10 as the backbone. The '*' indicates the default value used in all other experiments.
	}
	\label{fig:hyperparameters}
\end{figure}

\textbf{Effect of loss components.} We measure the individual contribution of $L_\text{rank}$ and $L_\text{cl}$ in \cref{eq:loss_total} by deactivating each of them. Furthermore, we propose to replace $L_\text{rank}$ with standard cross-entropy losses to validate the benefit of our ranking and hinge loss formulation introduced in \cref{eq:loss_rank}. Specifically, we evaluate the following alternative objective functions:

\begin{subequations}\label{eq:Lrank_to_CE}
	\begin{align}
		L_\text{CE}^\text{att/amp}          & = \text{CE}(p_\text{att}^\text{adv}, y) - \text{CE}(p_\text{amp}, y), \label{eq:Lrank_to_CE_dual} \\
		L_\text{CE}^\text{att} \hspace{1em} & = \hspace{3em} \text{CE}(p_\text{att}^\text{adv}, y), \label{eq:Lrank_to_CE_att}                  \\
		L_\text{CE}^\text{amp} \hspace{1em} & = \hspace{2.25em} - \text{CE}(p_\text{amp}, y). \label{eq:Lrank_to_CE_amp}
	\end{align}
\end{subequations}

The results in \cref{tab:abl_loss} demonstrate that both $L_\text{rank}$ and $L_\text{cl}$ are essential for achieving high adversarial robustness. Moreover, replacing $L_\text{rank}$ with the unconstrained loss in \cref{eq:Lrank_to_CE_dual} leads to suboptimal performance, indicating that the hinge formulation is necessary to focus the optimization on samples where the error gap between the amplified and attenuated activations is large. Similarly, a performance drop is observed when using \cref{eq:Lrank_to_CE_att,eq:Lrank_to_CE_amp}, which removes the dual optimization process between the two scaling modes. These findings confirm that our hinge-based ranking formulation is crucial for maintaining a stable optimization process and achieving high robust accuracy.

\textbf{Effect of the hyperparameters.} We analyze the impact of the hyperparameters in the A3 module. We empirically found the best configuration with $\tau_\mathit{m} = 0.1$, $\tau_\text{cl} = 10$, $\lambda_\text{rank} = 1$ and $\lambda_\text{cl} = 5$. To justify this choice, we vary each parameter while keeping the others fixed, as illustrated in \cref{fig:hyperparameters}. Our observations indicate that the module is sensitive to the choice of $\lambda_\text{rank}$, which is expected since this hyperparameter controls the cross-entropy component. We also observe a high sensitivity to $\lambda_\text{cl}$ and $\tau_\text{cl}$, indicating that the contrastive logit loss plays an important role in improving robustness. Finally, the performance remains relatively stable for $\tau_\mathit{m}$ when its value is below 1.

\begin{table}[!tb]
	\centering
	\caption{
		\textbf{Robust accuracy (\%) of A3 with different formulations of $f_m$/$f_{1m}$.} We use ResNet-18 trained with AT on CIFAR-10 as the backbone. The highlighted row corresponds to our default formulation presented in \cref{eq:fm}. Best results are in \textbf{bold}.
	}
	\setlength{\tabcolsep}{5pt}
	\resizebox{0.95\linewidth}{!}{
		\begin{tabular}{
			c
			>{\centering\arraybackslash}p{2.0cm}  
			>{\centering\arraybackslash}p{2.0cm}  
			|c|cccccc}
			\hline
			\centering \textbf{Method}            & $\boldsymbol{f_m}$                      & $\boldsymbol{f_{1m}}$               & \textbf{Clean}     & \textbf{FGSM}      & \textbf{PGD-20}    & \textbf{PGD-100}   & \textbf{C\&W}      & \textbf{Ens.}      & \textbf{AA}        \\
			\hline
			linear                                & $\frac{s_2}{s_1}$                         & $1 - f_{m}$                         & 84.12              & 57.94              & 51.95              & 49.56              & 51.22              & 48.58              & 46.85              \\
			linear (inverted)                     & $1 - f_{1m}$                            & $\frac{s_2}{s_1}$                     & 84.06              & 57.64              & 51.50              & 49.39              & 50.97              & 48.43              & 46.67              \\
			\hline
			quadratic                             & $(\frac{s_2}{s_1})^2$                     & $1 - f_{m}$                         & 84.34              & 59.70              & 54.93              & 52.92              & 51.27              & 49.41              & 46.71              \\
			quadratic (inverted)                  & $1 - f_{1m}$                            & $(\frac{s_2}{s_1})^2$                 & 84.22              & 57.71              & 51.75              & 49.69              & 51.01              & 48.80              & 46.22              \\
			\hline
			exponential                           & $\exp(1 - \frac{s_1}{s_2})$               & $1 - f_{m}$                         & 84.33              & 59.00              & 52.73              & 50.96              & 51.16              & 48.89              & 46.86              \\
			exponential (inverted)                & $1 - f_{1m}$                            & $\exp(1 - \frac{s_1}{s_2})$           & 84.26              & 57.72              & 51.86              & 50.05              & 50.79              & 48.57              & 46.13              \\
			\hline
			\CC log (\textbf{Ours}, \cref{eq:fm}) & \CC $\frac{\ln(1 + s_2)}{\ln(1 + s_1)}$ & \CC $1 - f_{m}$                     & \CC \textbf{84.37} & \CC \textbf{62.43} & \CC \textbf{58.66} & \CC \textbf{57.01} & \CC \textbf{52.33} & \CC \textbf{51.04} & \CC \textbf{47.28} \\
			log (inverted)                        & $ 1 - f_{1m}$                           & $\frac{\ln(1 + s_2)}{\ln(1 + s_1)}$ & 84.26              & 61.94              & 58.16              & 56.44              & 51.61              & 50.52              & 46.88              \\
			\hline
		\end{tabular}
	}
	\label{tab:abl_fm}
\end{table}
\begin{table}[tb]
	\centering
	\caption{
		\textbf{Computational efficiency of A3.} We compare the number of parameters (M) and FLOPs (G) with and without A3 module. For comparison, we also include the results of FSR \cite{Adv_FSR} and FTA2C \cite{Adv_FTA2C} which also use feature masking methods.
	}
	\setlength{\tabcolsep}{6pt}
	\resizebox{0.85\linewidth}{!}{
		\begin{tabular}{p{2.6cm}|cc|cc}
			\hline

			\multicolumn{1}{c|}{}      & \multicolumn{2}{c|}{\textbf{ResNet-18}} & \multicolumn{2}{c}{\textbf{WideResNet-34-10}}                                                              \\
			\hline
			\centering \textbf{Method} & \textbf{\# Params (M)}                  & \textbf{FLOPs (G)}                            & \textbf{\# Params (M)}     & \textbf{FLOPs (G)}            \\
			\hline
			\hline
			Vanilla (Baseline)         & 11.17                                   & 0.5567                                        & 46.16                      & 6.6647                        \\
			+ FSR \cite{Adv_FSR}       & 12.43 (+1.26)                           & 0.5768 (+0.0201)                              & 47.72 (+1.56)              & 6.7640 (+0.0993)              \\
			+ FTA2C \cite{Adv_FTA2C}   & 21.88 (+10.71)                          & 0.7278 (+0.1711)                              & 57.20 (+11.04)             & 7.2360 (+0.6713)              \\
			\CC + A3 \textbf{(Ours)}   & \CC 11.44 \textbf{(+0.27)}              & \CC 0.5588 \textbf{(+0.0021)}                 & \CC 46.57 \textbf{(+0.41)} & \CC 6.6738 \textbf{(+0.0091)} \\
			\hline
		\end{tabular}
	}
	\label{tab:results_flops}
\end{table}

\textbf{Choice of the scaling factors.} We empirically verify the effectiveness of the factors $f_m$ and $f_{1m}$ in \cref{eq:fm} by evaluating several alternative formulations, including linear, quadratic, and exponential functions. We also evaluate the configurations where the roles of $f_m$ and $f_{1m}$ are inverted. Detailed mathematical formulations are provided in \cref{tab:abl_fm}. It is important to note that these functions are constrained to the range $[0, 1]$ to preserve the intended behavior of the scaling operations in \cref{eq:zatt_zamp}. The results in \cref{tab:abl_fm} show that our proposed logarithmic formulation significantly outperforms the other formulations.

\textbf{Computational Efficiency.} We analyze the computational overhead introduced by A3 in \cref{tab:results_flops}. We report the number of parameters and FLOPs for ResNet-18 and WideResNet-34-10. The FLOPs are calculated using an input size of $3 \times 32 \times 32$ (CIFAR-10) with a batch size of 1. The results show that A3 introduces negligible overhead in terms of both parameters and FLOPs compared to other masking-based methods, while achieving higher robust accuracy.

\subsection{Activation Distribution Analysis}

\begin{figure}[tb]
	\centering
	\includegraphics[width=0.87\linewidth]{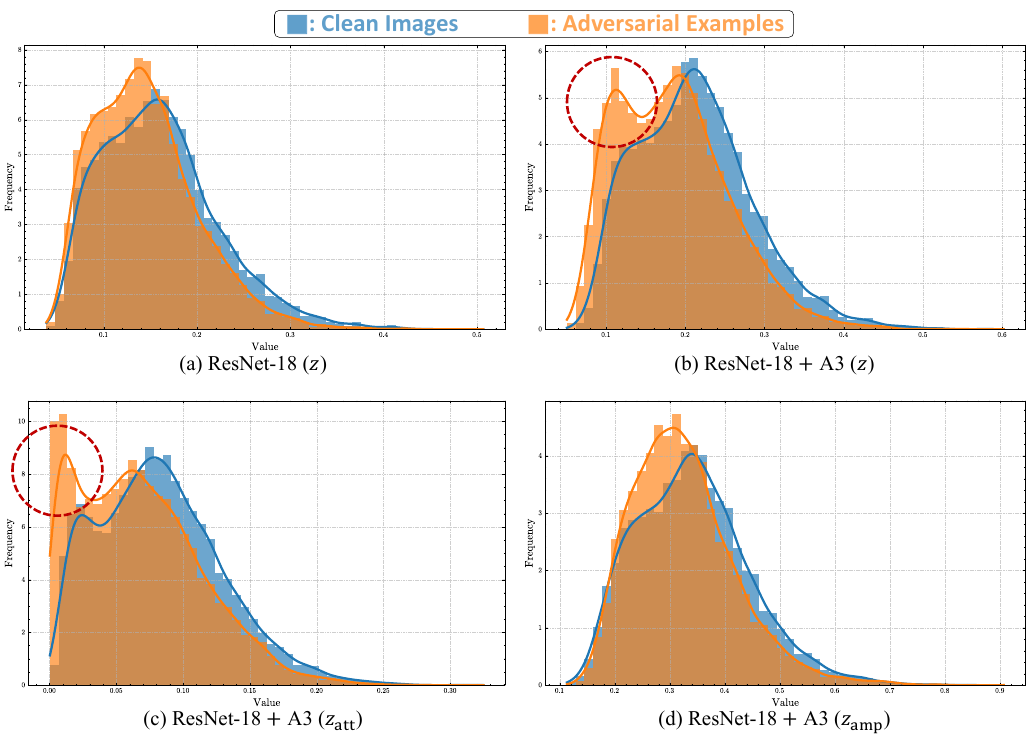}
	\caption{
		\textbf{Distribution of the activations under different modes.} We extract $z$ from ResNet-18 at \textit{block}~4 and visualize the distribution of average activation amplitudes for clean and adversarial examples. We compare four settings: (a)~ResNet-18 trained without A3, (b)~ResNet-18 trained with A3 (before scaling), (c)~A3 after attenuation ($z_\text{att}$), and (d)~A3 after amplification ($z_\text{amp}$).
		While clean and adversarial examples produce similar distributions in (a), the presence of A3 in (b) induces a clear distribution shift for adversarial examples (encircled in red). This shift is further accentuated in (c) via attenuation, thereby enhancing the separability between clean and adversarial examples. In contrast, the amplification mode in (d) increases the activation magnitudes for both inputs, resulting in overlapping distributions similar to (a).
	}
	\label{fig:act_res}
\end{figure}

To further understand how our A3 module improves adversarial robustness, we analyze the distribution of the activations before and after applying A3 to the backbone. \Cref{fig:act_res} presents the histogram of the averaged activation amplitudes collected from ResNet-18. In \cref{fig:act_res}~(a), we show the activations from a vanilla ResNet-18 after \textit{block}~4, \ie, where A3 is usually inserted. \Cref{fig:act_res}~(b) shows the activations from the same location in a model trained with A3. Interestingly, a distribution shift is already visible, indicating that the presence of the A3 module influences the learned representations during training. Specifically, the distribution in \cref{fig:act_res} (b) shows two distinct peaks, suggesting that the network has already learned to separate the activations into two groups. The left peak (encircled in red) only appears with adversarial examples. We hypothesize that these activations correspond to activation patterns induced by adversarial perturbations, which are effectively suppressed by A3.

We show the activations frequency distribution after applying the A3 module in attenuation mode ($z_\text{att}$) in \cref{fig:act_res} (c). Here, the left peak originally observed in \cref{fig:act_res} (b) is effectively suppressed for adversarial examples, bringing it close to zero. The effect is less pronounced for clean images, which is consistent with our previous analysis in \cref{tab:results_resnet18} that clean accuracy remains relatively unaffected while adversarial robustness improves.

In contrast, the amplification mode ($z_\text{amp}$) in \cref{fig:act_res} (d) shows substantial overlap between the clean and adversarial distributions, similar to the vanilla model in \cref{fig:act_res} (a). This indicates that while the attenuation mode enhances separability between clean and adversarial examples, the amplification mode intentionally reduces this separability, leading to the performance degradation observed in \cref{tab:abl_att_vs_amp}.

\section{Conclusion}

In this work, we introduced \textbf{Activation Amplification and Attenuation (A3)}, a lightweight plug-in module designed to improve adversarial robustness through controlled rescaling operations on the intermediate activations. A3 computes a scalar factor based on the original activation magnitudes to selectively amplify or attenuate intermediate activation patterns. The key idea is to use these amplified and attenuated activations during training to jointly learn robustness enhancement and prediction degradation. This process is optimized using novel ranking and contrastive loss functions that leverage the amplified activations as a negative reference. Extensive experiments demonstrate that A3 consistently improves adversarial robustness across different models, datasets, and training strategies while introducing negligible computational and memory overhead. By offering an efficient feature-based defense with few trainable parameters, A3 provides a promising direction for building more robust and interpretable models.

\section*{Acknowledgements}

This work was supported by JSPS KAKENHI (Grant Numbers JP24K22315 and JP26K21240) and JST BOOST (Grant Number JPMJBS2421).

%
%
\bibliographystyle{splncs04}
\bibliography{main}

\clearpage
\appendix


\section{Additional Results}

In this section, we provide additional experimental results complementing those presented in the main paper.

\textbf{Confidence Score.} To further understand the behavior of the A3 module, we analyze the confidence scores predicted by the model on clean and adversarial examples. Although softmax confidence scores do not necessarily provide accurate uncertainty estimates \cite{OOD_Energy}, they remain a useful diagnostic metric to compare the overall behavior of different models. We show in \cref{tab:analysis_conf_conf} the average confidence scores for the correct and incorrect predictions under different attacks. As expected, all models show higher confidence on correct predictions than on incorrect ones. However, the baseline ResNet-18 yields high confidence for incorrect predictions under adversarial attacks (\eg 63.77\% under PGD-100), indicating overconfident failures. The use of the A3 module in amplification mode ($z_\text{amp}$) increases the confidence scores (\eg, 87.23\% under PGD-100 on incorrect predictions), while the attenuation mode ($z_\text{att}$) decreases the confidence on errors (\eg, 43.78\% under PGD-100). This indicates that the attenuation mode of A3 helps mitigate overconfident predictions on adversarial examples, thus improving the model's reliability.

\textbf{Expected Calibration Error.} We compute the Expected Calibration Error (ECE)~\cite{ECE} to better quantify the alignment between confidence scores and accuracy of the models. Let $\hat{p}_i \in [0, 1]$ be the confidence of sample $i$, and let $\hat{y}_i$ and $y_i$ be the predicted and ground-truth labels, respectively. We partition $[0, 1]$ into $M$ equal-width bins $\{B_m\}_{m=1}^M$, and compute the accuracy and confidence for each bin as follows:

\begin{equation}
	\text{acc}(B_m) = \frac{1}{|B_m|} \sum_{i \in B_m} \mathbbm{1}(\hat{y}_i = y_i), \qquad
	\text{conf}(B_m) = \frac{1}{|B_m|} \sum_{i \in B_m} \hat{p}_i,
\end{equation}

Then, the ECE is obtained by computing the weighted average of the absolute difference between accuracy and confidence across all bins:

\begin{equation}
	\text{ECE} = \sum_{m=1}^{M} \frac{|B_m|}{n} |\text{acc}(B_m) - \text{conf}(B_m)|.
\end{equation}

A lower ECE indicates better calibration of the confidence scores. The results in \cref{tab:analysis_conf_ece} show that the attenuation mode ($z_\text{att}$) consistently reduces ECE compared to the baseline, while the amplification mode ($z_\text{amp}$) leads to higher ECE. These results further support the observation that the attenuation mode of A3 reduces model overconfidence.

\begin{table}[!tb]
	\centering
	\caption{
		\textbf{Confidence score analysis.} We report the confidence scores of ResNet-18 with A3 in amplification ($z_\text{amp}$) and attenuation ($z_\text{att}$) modes. The confidence score is defined as the maximum softmax probability. In \cref{tab:analysis_conf_conf}, we separate the average confidence scores for correct and incorrect predictions. In \cref{tab:analysis_conf_ece}, we report the Expected Calibration Error (ECE) scores computed with 15 equal-width bins. We do not highlight "best" values in \cref{tab:analysis_conf_conf} since raw confidence has no unique optimum (\eg, higher confidence can correspond to overconfident errors depending on the attack strength and image clarity \cite{OOD_Energy}). In contrast, ECE provides a quantitative measure of calibration quality, where lower values indicate better calibration.
	}
	\begin{subtable}[t]{\linewidth}
		\centering
		\caption{Confidence Score (correct / incorrect)}
		\setlength{\tabcolsep}{5pt}
		\resizebox{0.95\linewidth}{!}{
			\begin{tabular}{
					p{4.2cm}
					|c|c|c|c|c|c}
				\hline
				\multicolumn{1}{c|}{\textbf{Method}}      & \multicolumn{1}{c}{\textbf{Clean}} & \multicolumn{1}{c}{\textbf{FGSM}} & \multicolumn{1}{c}{\textbf{PGD-20}} & \multicolumn{1}{c}{\textbf{PGD-100}} & \multicolumn{1}{c}{\textbf{C\&W}} & \multicolumn{1}{c}{\textbf{Average}} \\
				\hline
				ResNet-18 (Baseline)                       & 84.35 / 50.21                      & 79.69 / 57.96                     & 80.08 / 61.91                       & 80.66 / 63.77                        & 82.27 / 55.21                     & 81.41 / 57.81                        \\
				+ A3 | Amplification ($z_\text{amp}$)   & 94.72 / 72.44                      & 92.50 / 82.00                     & 92.66 / 85.82                       & 92.76 / 87.23                        & 92.71 / 77.30                     & 93.07 / 80.96                        \\
				\CC + A3 | Attenuation ($z_\text{att}$) & \CC 71.49 / 35.08                  & \CC 62.17 / 38.04                 & \CC 60.94 / 41.65                   & \CC 60.87 / 43.78                    & \CC 67.69 / 38.55                 & \CC 64.63 / 39.42                    \\
				\hline
			\end{tabular}
			\label{tab:analysis_conf_conf}
		}
	\end{subtable}

	\vspace{0.3em}

	\begin{subtable}[t]{\linewidth}
		\centering
		\caption{ECE Score (lower is better)}
		\setlength{\tabcolsep}{10pt}
		\resizebox{0.95\linewidth}{!}{
			\begin{tabular}{
					p{4.2cm}
					|cccccc}
				\hline
				\multicolumn{1}{c|}{\textbf{Method}}      & \multicolumn{1}{c}{\textbf{Clean}} & \multicolumn{1}{c}{\textbf{FGSM}} & \multicolumn{1}{c}{\textbf{PGD-20}} & \multicolumn{1}{c}{\textbf{PGD-100}} & \multicolumn{1}{c}{\textbf{C\&W}} & \multicolumn{1}{c}{\textbf{Average}} \\
				\hline
				ResNet-18 (Baseline)                       & 0.0715                             & 0.1300                            & 0.1755                              & 0.2003                               & 0.1538                            & 0.1462                               \\
				+ A3 | Amplification ($z_\text{amp}$)   & 0.1516                             & 0.2601                            & 0.3184                              & 0.3391                               & 0.2759                            & 0.2690                               \\
				\CC + A3 | Attenuation ($z_\text{att}$) & \CC \textbf{0.0685}                & \CC \textbf{0.1224}               & \CC \textbf{0.1372}                 & \CC \textbf{0.1404}                  & \CC \textbf{0.0875}               & \CC \textbf{0.1112}                  \\
				\hline
			\end{tabular}
			\label{tab:analysis_conf_ece}
		}
	\end{subtable}

	\label{tab:analysis_conf}
\end{table}

\section{Additional Ablation Studies}

To better understand the behavior of A3, we conduct additional ablation studies that are not included in the main paper.

\textbf{Effect of module position.} Although we mainly insert the A3 module after the last block of the backbone network in our main experiments (\ie, \textit{block}~3 for WideResNet-34-10 and \textit{block}~4 for ResNet-18), it is possible to place the module at different locations within the network. We conduct an ablation study using ResNet-18 and WideResNet-34-10 trained on CIFAR-10 with standard adversarial training (AT) to evaluate the impact of the module position. The results in \cref{tab:abl_block} show that inserting A3 in earlier blocks leads to lower robustness compared to our default configuration. This suggests that applying A3 too early in the network may interfere with the acquisition of low-level features. Furthermore, the limited number of channels in earlier blocks may limit the capacity of the module to generate meaningful masks. This trend aligns with the observations from existing feature-based defense methods~\cite{Adv_FSR,Adv_FTA2C}, which also place their modules in the deeper layers of the network.

\begin{table}[t]
	\centering
	\caption{
		\textbf{Robust accuracy (\%) when inserting A3 module at different blocks.} We trained the models with AT on CIFAR-10 using ResNet-18 and WideResNet-34-10. The highlighted row corresponds to our default setting for all other experiments. The best results are highlighted in \textbf{bold}.
	}
	\setlength{\tabcolsep}{3pt}

	\vspace{-1.0em}

	\begin{subtable}[t]{\linewidth}
		\centering
		\caption{ResNet-18}
		\resizebox{0.95\linewidth}{!}{
			\begin{tabular}{p{3.1cm}|c|cccccc}
				\hline
				\centering \textbf{Method}           & \textbf{Clean} & \textbf{FGSM}      & \textbf{PGD-20}    & \textbf{PGD-100}   & \textbf{C\&W}      & \textbf{Ens.}      & \textbf{AA}        \\
				\hline
				ResNet-18 (Baseline)                 & 85.04          & 56.96              & 49.12              & 47.51              & 48.19              & 46.05              & 44.28              \\
				+ A3 (\textit{block}~1)              & \textbf{85.27} & 58.22              & 52.12              & 49.48              & 51.40              & 48.58              & 47.02              \\
				+ A3 (\textit{block}~2)              & 84.61          & 57.35              & 51.19              & 48.99              & 50.38              & 47.71              & 46.06              \\
				+ A3 (\textit{block}~3)              & 79.11          & 55.34              & 51.07              & 49.62              & 48.61              & 47.85              & 44.75              \\
				\CC \textbf{+ A3 (\textit{block}~4)} & \CC 84.37      & \CC \textbf{62.43} & \CC \textbf{58.66} & \CC \textbf{57.01} & \CC \textbf{52.33} & \CC \textbf{51.04} & \CC \textbf{47.28} \\
				+ A3 (\textit{block}~3+4)            & 71.13          & 47.18              & 44.28              & 43.31              & 39.18              & 38.47              & 36.87              \\
				\hline
			\end{tabular}
		}
	\end{subtable}

	\vspace{0.3em}

	\begin{subtable}[t]{\linewidth}
		\centering
		\caption{WideResNet-34-10}
		\resizebox{0.95\linewidth}{!}{
			\begin{tabular}{p{4.2cm}|c|cccccc}
				\hline
				\centering \textbf{Method}           & \textbf{Clean} & \textbf{FGSM}      & \textbf{PGD-20}    & \textbf{PGD-100}   & \textbf{C\&W}      & \textbf{Ens.}      & \textbf{AA}        \\
				\hline
				WideResNet-34-10 (Baseline)          & \textbf{87.57} & 60.12              & 51.58              & 49.83              & 51.60              & 49.27              & 48.18              \\
				+ A3 (\textit{block}~1)              & 77.46          & 51.22              & 47.83              & 46.55              & 45.01              & 43.93              & 42.55              \\
				+ A3 (\textit{block}~2)              & 87.34          & 61.61              & 55.70              & 53.76              & 55.17              & 53.02              & 51.69              \\
				\CC \textbf{+ A3 (\textit{block}~3)} & \CC 86.64      & \CC \textbf{63.01} & \CC \textbf{58.33} & \CC \textbf{56.72} & \CC \textbf{55.32} & \CC \textbf{53.82} & \CC \textbf{51.62} \\
				\hline
			\end{tabular}
		}
	\end{subtable}

	\label{tab:abl_block}
\end{table}
\begin{table}[!tb]
	\centering
	\caption{
		\textbf{Robust accuracy (\%) of A3 using different masking functions.} We use ResNet-18 and WideResNet-34-10 trained with AT on CIFAR-10 as the backbone. We compare our default Gumbel-Softmax mask with a hard-threshold binary mask as defined in \cref{eq:hard_threshold}. Both methods improve robustness over the baseline, but with significantly higher performance using the Gumbel-Softmax method. The best results are highlighted in \textbf{bold}.
	}
	\setlength{\tabcolsep}{4pt}
	\resizebox{0.95\linewidth}{!}{
		\begin{tabular}{
				p{4.2cm}
				|c|cccccc}
			\hline
			\centering \textbf{Method}  & \textbf{Clean} & \textbf{FGSM}      & \textbf{PGD-20}    & \textbf{PGD-100}   & \textbf{C\&W}      & \textbf{Ens.}      & \textbf{AA}        \\
			\hline
			ResNet-18 (Baseline)         & \textbf{85.04} & 56.96              & 49.12              & 47.51              & 48.19              & 46.05              & 44.28              \\
			+ A3 (Binary)               & 83.93          & 58.24              & 52.23              & 50.10              & 51.28              & 48.79              & 46.83              \\
			\CC \textbf{+ A3 (Gumbel)}  & \CC 84.37      & \CC \textbf{62.43} & \CC \textbf{58.66} & \CC \textbf{57.01} & \CC \textbf{52.33} & \CC \textbf{51.04} & \CC \textbf{47.28} \\
			\hline
			WideResNet-34-10 (Baseline) & \textbf{87.57} & 60.12              & 51.58              & 49.83              & 51.60              & 49.27              & 48.18              \\
			+ A3 (Binary)               & 86.54          & 60.76              & 55.00              & 52.18              & 54.33              & 51.31              & 49.82              \\
			\CC \textbf{+ A3 (Gumbel)}  & \CC 86.64      & \CC \textbf{63.01} & \CC \textbf{58.33} & \CC \textbf{56.72} & \CC \textbf{55.32} & \CC \textbf{53.82} & \CC \textbf{51.62} \\
			\hline
		\end{tabular}
	}
	\label{tab:abl_mask}
\end{table}

\textbf{Choice of the masking function.} The use of the Gumbel-Softmax function in \cref{eq:GumbelSoftmax} is primarily motivated by its differentiability, as also highlighted in prior works such as FSR~\cite{Adv_FSR} and FTA2C~\cite{Adv_FTA2C}. To validate this design choice, we conduct an ablation study where we replace the Gumbel-Softmax trick with a hard-threshold mask defined as follows:

\begin{equation}\label{eq:hard_threshold}
	\text{Mask}_\text{hard}(z_m) =
	\begin{cases}
		1, & \text{if } z_m > 0 \\
		0, & \text{otherwise}
	\end{cases}.
\end{equation}

The results in \cref{tab:abl_mask} show that training A3 with a hard-threshold mask still yields consistent robustness gains over the vanilla backbone. This indicates that A3's improvements do not rely solely on the Gumbel-Softmax operator, and that the underlying scaling mechanism is beneficial in itself for adversarial robustness. However, the Gumbel-Softmax variant achieves higher robust accuracy, suggesting that mask differentiability facilitates gradient-based, end-to-end optimization of the masking and scaling behavior. These findings are consistent with the observations in FSR~\cite{Adv_FSR}, which similarly report a performance drop when replacing their Gumbel-Softmax mask with a hard-threshold alternative.

\begin{figure}[tb]
	\centering
	\includegraphics[width=0.95\linewidth]{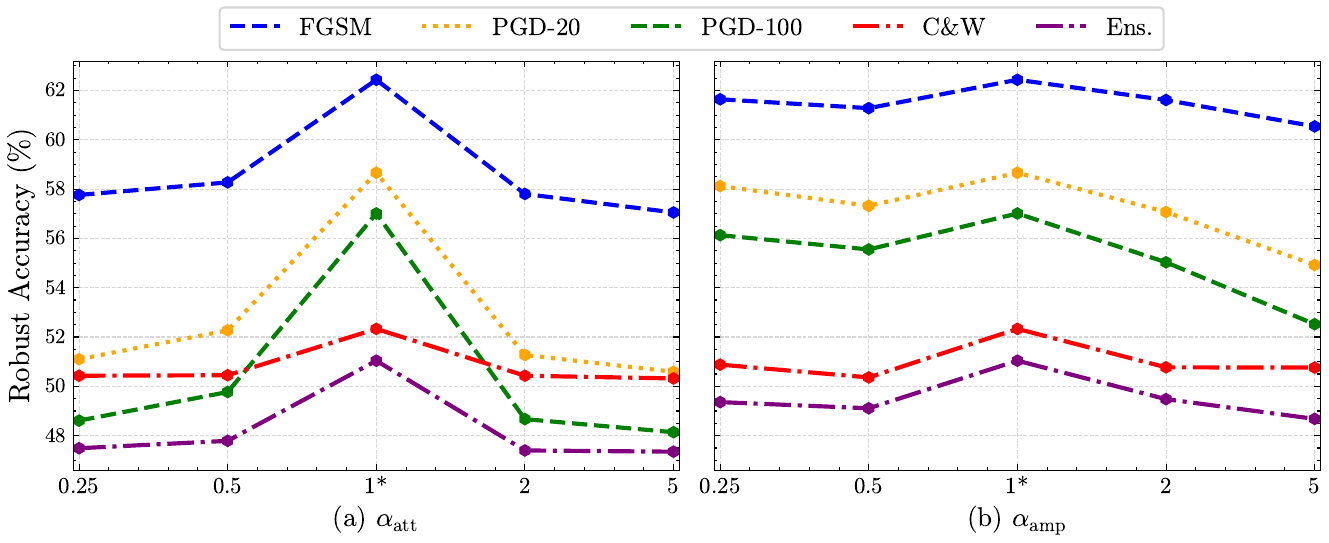}
	\caption{
		\textbf{Effect of the scaling intensity.} We adjust the intensity of the scaling operation in \cref{eq:zatt_zamp_alpha} with a constant: (a) $\alpha_\text{amp}$ for the amplification mode and (b) $\alpha_\text{att}$ for the attenuation mode. All other experiments use $\alpha_\text{amp} = \alpha_\text{att} = 1$ (marked by '*').
	}
	\label{fig:ablation_alphas}
\end{figure}

\textbf{Effect of the scaling intensity.} In the main paper, we show that the amplification mode significantly enhances the performance of the attenuation mode, thus improving overall robustness. A question that naturally arises is whether A3 can be further optimized by increasing the intensity of the amplification mode. To evaluate this, we introduce two hyperparameters $\alpha_\text{att}$ and $\alpha_\text{amp}$ to modulate the intensity of the amplification and attenuation operations in \cref{eq:zatt_zamp} as follows:

\begin{equation}\label{eq:zatt_zamp_alpha}
	z_{\text{att}} = z \cdot \alpha_\text{att} \cdot \bigl[1 - \text{Scale}(z,m) \bigr], \qquad
	z_{\text{amp}} = z \cdot \alpha_\text{amp} \cdot \bigl[1 + \text{Scale}(z,m) \bigr].
\end{equation}

Our default configuration used in all other experiments is $\alpha_\text{att} = \alpha_\text{amp} = 1$. The results in \cref{fig:ablation_alphas} show that varying the intensity of either the amplification or attenuation mode leads to a decrease in robust accuracy. This suggests that our original formulation already provides a strong trade-off between the two modes. Specifically, we hypothesize that excessively large amplification may lead to severe distortions of the original activations, destabilizing the intended optimization process. Conversely, reducing the scaling intensity likely weakens the contrast between the amplified and attenuated activations, which reduces the effectiveness of the ranking and contrastive losses. Overall, these results indicate that the balance between the amplification and attenuation modes is crucial for achieving optimal robustness, and that simply increasing the strength of one mode does not necessarily lead to further improvements.

\section{Analysis on Obfuscated Gradients}

As discussed in~\cite{Adv_Adaptive,Adv_RobustBench,Adv_AutoAttack,Adv_Obfuscation}, some defense methods~\cite{kWTA,Adv_SAP} may rely on obfuscated gradients to achieve adversarial robustness. To verify that our proposed A3 module does not fall into this category, we follow the evaluation guidelines of~\cite{Adv_Obfuscation} and verify that our method satisfies the following three properties:

\textbf{(i) The attack performance increases with the number of iterations.} Methods such as FGSM~\cite{AdvAtt_Pandas} and \text{PGD-10/20/100}~\cite{AdvAtt_PGD} are based on the same algorithm, but with a different number of attack iterations per sample. As such, robust accuracy is expected to decrease as the number of attack iterations increases (\eg, PGD-100 should be stronger than PGD-20).

\begin{figure}[!tb]
	\centering
	\begin{subfigure}{0.49\linewidth}
		\includegraphics[width=\linewidth]{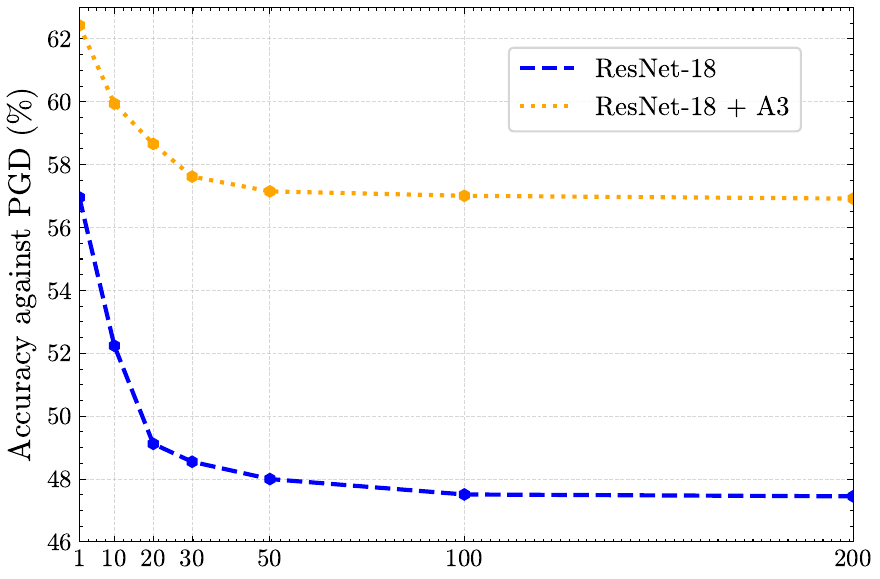}
		\caption{Number of iterations}
		\label{fig:obf_iter}
	\end{subfigure}
	\centering
	\begin{subfigure}{0.49\linewidth}
		\includegraphics[width=\linewidth]{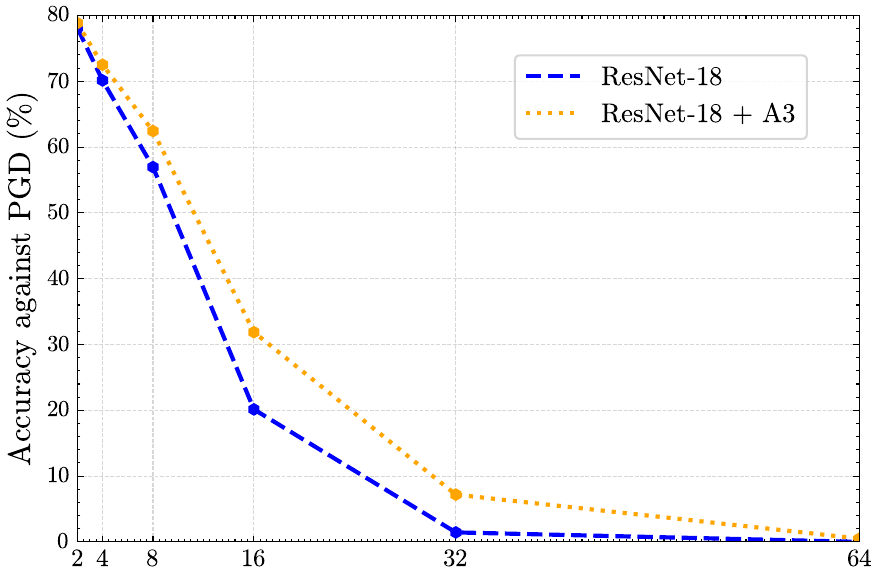}
		\caption{Perturbation budget $\epsilon$ ($\text{x} / 255$)}
		\label{fig:obf_eps}
	\end{subfigure}
	\caption{
		\textbf{Effect of attack strength on model robustness.} Using ResNet-18 trained with AT and AT + A3 on CIFAR-10, we compute the robust accuracy under PGD attacks with varying attack strength. In \cref{fig:obf_iter}, we vary the number of attack iterations from $1$ to $200$ while fixing the perturbation budget to $\epsilon=8/255$. In \cref{fig:obf_eps} we vary the perturbation budget $\epsilon$ from $2/255$ to $64/255$ while fixing the number of iterations to $20$. As expected, the robust accuracy decreases when increasing either $\epsilon$ or the number of iterations (\ie, when increasing the attack strength). Importantly, the gap between the two curves remains consistent across attack strengths, indicating that A3 provides stable robustness improvements rather than gains limited to specific settings. This behavior provides further evidence that A3 is not relying on gradient obfuscation.
	}
	\label{fig:obfuscated_gradients}
\end{figure}

\textbf{Results.} The results in \cref{tab:results_resnet18,tab:results_wideresnet3410,tab:results_tinyimagenet} are consistent with the expected behavior, as the robust accuracy under PGD-100 is consistently lower than under PGD-20 and FGSM for all the models trained with A3. In addition, we show in \cref{fig:obf_iter} the robust accuracy under PGD attacks with the number of iterations varying from 1 to 200. The curves for both ResNet-18 and ResNet-18 + A3 show a decreasing trend as the number of iterations increases, with a consistent gap between the two models. This indicates that A3 provides stable robustness improvements, which is consistent with the observations expected in the absence of gradient obfuscation.

\textbf{(ii) The attack performance increases with the attack strength.} In our work, we generate all the adversarial examples using the $l_{\infty}$ norm with $\epsilon=8/255$ as the perturbation budget, which is a common setting in the literature~\cite{Adv_RobustBench,Adv_AutoAttack}. It is natural to expect that the robust accuracy would decrease as the perturbation budget $\epsilon$ increases, until reaching a point where the accuracy becomes close to zero as the perturbations severely distort the input images. If this behavior is not observed, it may indicate that the model relies on randomness or other factors that hinder fair robustness evaluation.

\textbf{Results.} We show in \cref{fig:obf_eps} the robust accuracy under PGD-20 attacks with a perturbation budget $\epsilon$ varying from $2/255$ to $64/255$. The curves for both ResNet-18 and ResNet-18 + A3 show decreasing robust accuracy as the perturbation budget increases. Furthermore, the accuracy gap between the two models is larger around $\epsilon=16/255$. This is likely because very small perturbation budgets (\eg, $2/255$) are too weak to generate effective adversarial examples, leading to a high robust accuracy for both models with only a small performance gap. In contrast, very large budgets (\eg, $64/255$) heavily distort the input images, making the classification task extremely difficult for both models.

\begin{figure}[!tb]
	\centering
	\includegraphics[width=0.64\linewidth]{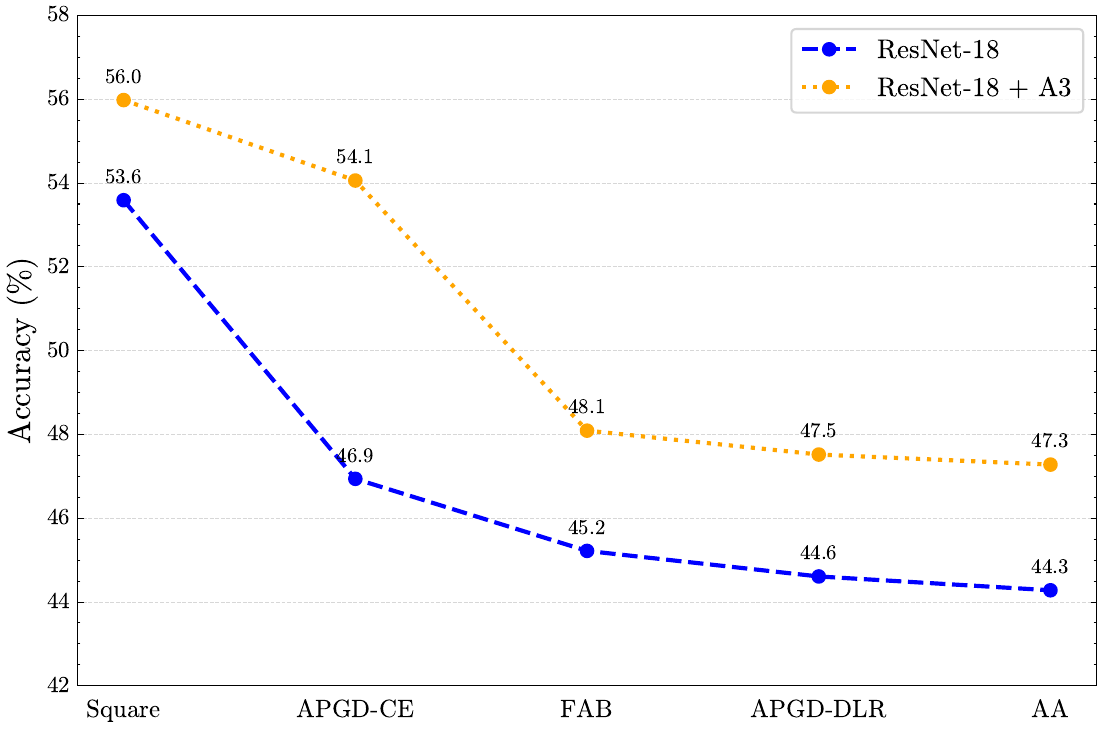}
	\caption{
		\textbf{Detailed comparison of AutoAttack results.} We report the robust accuracy (\%) under AutoAttack (AA) using ResNet-18 trained with AT on CIFAR-10. Since AA is an ensemble of four attacks (APGD-CE, APGD-DLR, FAB and Square), we also report the accuracy of each individual attack. As expected, AA yields the lowest accuracy as it corresponds to the sample-wise worst-case performance across the four attacks. Furthermore, Square attack yields the highest accuracy, which is also expected since Square is a black-box attack and is generally weaker than white-box attacks (APGD-CE, APGD-DLR and FAB).
	}
	\label{fig:autoattack_bar}
\end{figure}

\textbf{(iii) White-box attacks are stronger than black-box attacks.} In white-box attacks, the attacker has full access to the model parameters and gradients, while in black-box attacks, the attacker can only query the model outputs. Therefore, unless a specific circumstance exists (\eg, unreliable gradients due to obfuscation), white-box attacks are generally expected to be stronger than black-box attacks.

\textbf{Results.} To verify this property, we utilize AutoAttack (AA)~\cite{Adv_AutoAttack} as it is a widely adopted ensemble of attacks that includes both white-box and black-box attacks. In \cref{tab:results_resnet18,tab:results_wideresnet3410,tab:results_tinyimagenet}, obtaining a higher robust accuracy under AA compared to the baselines suggests a stronger robustness to worst-case scenarios. Furthermore, we report the individual accuracy of each attack included in AA (APGD-CE, APGD-DLR, FAB and Square). The results in \cref{fig:autoattack_bar} show that the black-box attack (Square) is weaker than all the white-box attacks (APGD-CE, APGD-DLR and FAB), which is consistent with the expected behavior.

\end{document}